\documentclass{article}

\usepackage{microtype}
\usepackage{graphicx}
\usepackage{subfigure}
\usepackage{booktabs} % for professional tables

\usepackage{hyperref}

\usepackage[accepted]{icml2025}

\usepackage{amsmath}
\usepackage{amssymb}
\usepackage{mathtools}
\usepackage{amsthm}

\usepackage[capitalize,noabbrev]{cleveref}

\usepackage{thmtools, thm-restate}
\theoremstyle{plain}
\newtheorem{theorem}{Theorem}[section]
\newtheorem{proposition}[theorem]{Proposition}

\theoremstyle{definition}
\newtheorem{definition}[theorem]{Definition}

\theoremstyle{remark}
\newtheorem{remark}[theorem]{Remark}

\usepackage[textsize=tiny]{todonotes}

\icmltitlerunning{Differentially Private Policy Gradient}

\begin{document}

\twocolumn[
\icmltitle{Differentially Private Policy Gradient}

% It is OKAY to include author information, even for blind
% submissions: the style file will automatically remove it for you
% unless you've provided the [accepted] option to the icml2025
% package.

% List of affiliations: The first argument should be a (short)
% identifier you will use later to specify author affiliations
% Academic affiliations should list Department, University, City, Region, Country
% Industry affiliations should list Company, City, Region, Country

% You can specify symbols, otherwise they are numbered in order.
% Ideally, you should not use this facility. Affiliations will be numbered
% in order of appearance and this is the preferred way.
%\icmlsetsymbol{equal}{*}

\begin{icmlauthorlist}
\icmlauthor{Alexandre Rio}{huawei}
\icmlauthor{Merwan Barlier}{huawei}
\icmlauthor{Igor Colin}{telecom}
%\icmlauthor{}{sch}
%\icmlauthor{}{sch}
\end{icmlauthorlist}

\icmlaffiliation{huawei}{Noah’s Ark Lab Paris, Huawei Technologies France, Paris}
\icmlaffiliation{telecom}{Télécom Paris, France}

\icmlcorrespondingauthor{Alexandre Rio}{alexandre.rio2@huawei.com}

% You may provide any keywords that you
% find helpful for describing your paper; these are used to populate
% the "keywords" metadata in the PDF but will not be shown in the document
\icmlkeywords{Machine Learning, ICML}

\vskip 0.3in
]

% this must go after the closing bracket ] following \twocolumn[ ...

% This command actually creates the footnote in the first column
% listing the affiliations and the copyright notice.
% The command takes one argument, which is text to display at the start of the footnote.
% The \icmlEqualContribution command is standard text for equal contribution.
% Remove it (just {}) if you do not need this facility.

%\printAffiliationsAndNotice{}  % leave blank if no need to mention equal contribution
\printAffiliationsAndNotice{\icmlEqualContribution} % otherwise use the standard text.

\begin{abstract}
    Motivated by the increasing deployment of reinforcement learning in the real world, involving a large consumption of personal data, we introduce a differentially private (DP) policy gradient algorithm. We show that, in this setting, the introduction of Differential Privacy can be reduced to the computation of appropriate trust regions, thus avoiding the sacrifice of theoretical properties of the DP-less methods. Therefore, we show that it is possible to find the right trade-off between privacy noise and trust-region size to obtain a performant differentially private policy gradient algorithm. We then outline its performance empirically on various benchmarks. Our results and the complexity of the tasks addressed represent a significant improvement over existing DP algorithms in online RL.
\end{abstract}

%%%%%%%%%%%%%%%%%%%%%%%%%%%%%%%%%%%

\section{Introduction}

In the past years, Reinforcement Learning (RL) agents have been increasingly deployed in the real world, especially to provide personalized recommendations, spanning various domains such as healthcare \cite{rl_treatment_recommendation}, recommendation engines \cite{rl_recommendation_system}, and finance \cite{rl_finance}, or to control advanced systems, for instance autonomous vehicles \cite{rl_autonomous_driving} and robotics \cite{rl_robotics}. RL has also greatly contributed to the huge success of Large Language Models (LLMs), being used to align these models with human preferences (\textit{i.e.}, Reinforcement Learning from Human Feedback (RLHF), see \citet{rlhf_survey}) and to enhance their reasoning capabilities \cite{deepseekr1}. To succeed, these RL agents are typically trained on a large amount of personal data. 

As a consequence, there has been growing concern over privacy leakage from RL agents. Indeed, empirical evidence suggest that RL is no more immune to privacy attacks than other areas of ML. In particular, \citet{gomrokchi_membership_2022} exploit temporal data correlation in RL to design powerful membership inference attacks (MIAs) against RL policies in the black-box setting, being able to infer the membership of full trajectories to the training data. \citet{pan_how_2019} and \citet{prakash_how_2022} have also designed privacy attacks against the transition dynamics and the reward function of the underlying Markov decision process (MDP). Moreover, several studies have shown evidence that privacy leakage can happen during the fine-tuning training phase of LLMs \cite{CarliniTWJHLRBS21, Fu_FineTuningAttack2024}, suggesting that RLHF on private data may be unsafe. The existence of such privacy threats is a huge issue as the targeted data can be highly sensitive in some scenarios. For instance, in RL-based personalized treatment recommendation, states and rewards can reveal information about a patient's health history, while systems for online advertising or news recommendation are trained on users' browsing history that can reveal the users' preferences on various topics (\textit{e.g.}, politics). 

Designing privacy-preserving reinforcement learning methods, notably based on differential privacy (DP) --- the current gold standard in terms of privacy protection ---- is therefore of great importance. Yet, private reinforcement learning has received little attention compared to other areas of machine learning. A handful of works have introduced DP RL methods (\textit{e.g.}, \citet{vietri2020private}, \citet{chowdhury_differentially_2021}, \citet{qiao_tabular_2023}), but these works remain largely theoretical. Indeed, they are limited to episodic tabular and linear MDPs, only evaluated on low-dimensional simulations, and cannot, intrinsically, scale to real-world problems. Crucially, no work has convincingly tackled deep RL problems with differential privacy guarantees, even in relatively low-dimensional settings, and there currently is no DP RL method matching the versatility, scalability and empirical effectiveness of \textsc{DP-SGD} \cite{Abadi_2016} for (deep) supervised learning.

We therefore believe that there is a pressing need to steer the DP RL field towards more practical, scalable approaches, which starts by tackling deep RL problems with differential privacy guarantees. We further observe that policy gradient methods \cite{Policy_Gradient_SuttonMSM99}, although widely popular and achieving state-of-the-art results on various deep RL tasks, have not been studied in the private setting. Yet, directly optimizing the policy through gradient ascent, policy gradient methods exhibit clear similarities with classic deep learning optimization techniques like SGD. This similarity prompts exploring the use of clipped noisy gradient updates, which are at the core of \textsc{DP-SGD}, to derive differentially private versions of policy gradient algorithms. Moreover, policy gradient methods are at the core of RLHF for LLM alignment: given the current popularity of the topic and the associated privacy concerns, researching DP policy gradient methods seems highly warranted. 

Further motivated by the apparent connection between gradient clipping in DP-SGD and trust-region approaches in RL, we investigate the subclass of policy gradient methods referred to as \textit{trust-region methods}, which includes \textsc{TRPO} \cite{TRPOSchulmanLAJM15} and \textsc{PPO} \cite{PPOSchulmanWDRK17}. These algorithms, extensively used for their stability and versatility, aim at preventing too large policy updates which can result in dramatic performance drop during training. In the presence of DP noise, we characterize the inclusion of the gradient updates within a given trust region in terms of the Fisher Information matrix, and derive theoretical insights into how to set the DP clipping norm for the noisy updates to remain within the trust region with high probability, which we validate empirically on a discrete environment. Relaxing the theory, we then provide a practical PPO-like algorithm with formal, trajectory-level differential privacy guarantees. %Our approach achieves significant differential privacy guarantees with limited performance cost on previously unaddressed deep RL benchmarks, 
For realistic privacy budgets, our approach shows limited impact on the performance of policy gradient on previously unaddressed deep RL benchmarks, effectively paving the way for deploying DP RL agents in complex environments.
\section{Related Work}

\paragraph{Policy Gradient Methods in RL.} In RL, policy gradient (PG) methods \cite{Policy_Gradient_SuttonMSM99} optimize the policy directly within its parameter space using the policy gradient theorem. This theorem enables gradient ascent by formulating the gradient of the expected discounted return with respect to the policy parameters. Many policy gradient algorithms have been proposed over the years to improve upon the vanilla, high-variance \textsc{Reinforce} \cite{REINFORCEWilliams92}, among which \textit{actor-critic methods} \cite{A3CMnihBMGLHSK16, SAC_HaarnojaZAL18}, and \textit{deterministic PG methods} \cite{DPGSilverLHDWR14, DDPGLillicrapHPHETS15}. In this work, we focus on \textit{trust-region methods} which constrain policy updates while optimizing a surrogate loss. \textsc{TRPO} \cite{TRPOSchulmanLAJM15}, of which Natural Policy Gradient \cite{NPGKakade} can be considered a special case, imposes a hard constraint on the KL divergence between consecutive policies. Its successor, \textsc{PPO} \cite{PPOSchulmanWDRK17}, uses a relaxed constraint and has been widely popular for its ease of implementation and state-of-the-art performance on various tasks. Recently, PG methods like \textsc{PPO} have been successfully applied to aligning LLMs with human preferences through RLHF \cite{rlhf_survey}.

\paragraph{Private RL.} Differential Privacy (DP, \citet{Dwork06}), the current gold standard in terms of privacy protection, has been the focus of extensive research work over the past two decades.
%, aiming for better privacy-utility trade-offs and tighter privacy analysis \cite{DworkRV10, DworkR16, mironov_renyi_2017}.
In particular, the ideas behind \textsc{DP-SGD} \cite{Abadi_2016} have made possible the training of deep neural networks with DP guarantees, encouraging the deployment of private models at scale, although many practical challenges remain (see \citet{ponomareva_how_2023}). Despite the growing deployment of RL in the real-world \cite{rl_treatment_recommendation, rl_recommendation_system, rl_autonomous_driving} and well-documented privacy threats \cite{pan_how_2019, prakash_how_2022, gomrokchi_membership_2022}, the DP RL literature remains mainly theoretical, limited to episodic tabular and linear MDPs (\textit{e.g.}, \citet{vietri2020private, chowdhury_differentially_2021, qiao_tabular_2023})
%\cite{vietri2020private, GarcelonPPP21, Liao2021, luyo2021differentially, chowdhury_differentially_2021, zhou_differentially_2022, ngo_linear_2022, qiao_offline_2022, qiao_tabular_2023}.
The proposed methods are fundamentally unable to scale to the problems encountered in the real world. A couple of works \cite{Wang2019, mutual_info_rl_2024} have addressed more general settings and harder problems, albeit with specific notions of privacy that differ from DP. To the best of our knowledge, no work has tackled deep RL problems with DP guarantees, and DP policy gradient methods remain unexplored.

% S'auto-citer?
\section{Background}

\subsection{Reinforcement Learning}

We consider a discounted infinite-horizon MDP $\mathcal{M} = \left(\mathcal{S}, \mathcal{A}, P, r, \rho_0, \gamma\right)$: $\mathcal{S}$ and $\mathcal{A}$ are the state and action spaces, respectively, $P:\mathcal{S} \times \mathcal{A} \times \mathcal{S} \rightarrow [0,1]$ is the transition probability function, $r:\mathcal{S}\times\mathcal{A} \rightarrow [0,1]$ is the reward function, $\rho_0: \mathcal{S} \rightarrow [0,1]$ is the starting state distribution and $\gamma \in (0,1)$ is a discount factor.
%In the sequel, $\mathcal{S}$ may be discrete or continuous, while $\mathcal{A}$ is discrete.
The goal of RL is to learn a stochastic policy $\pi: \mathcal{S} \times \mathcal{A} \rightarrow [0,1]$ which maximizes the expected discounted return:
\[
    J(\pi) = \mathbb{E}_{\rho_0, \pi, P}\left[\sum_{t \ge 0} \gamma^t r(s_t, a_t)\right] \enspace ,
\]
where the expectation subscripts indicate that $s_0 \sim\rho_0, a_t \sim \pi(\cdot \vert s_t)$ and $s_{t+1} \sim P(\cdot \vert s_t, a_t)$. We define the discounted return from step $t$ as $G_t = \sum_{k \ge 0} \gamma^k r(s_{t+k}, a_{t+k})$, the value function as $V^\pi(s) = \mathbb{E}_{\pi, P} \left[G_t\vert s_t = s\right]$, the state-action value function $Q^\pi(s,a) = \mathbb{E}_{\pi, P} \left[G_t \vert s_t=s,a_t=a\right]$ and the advantage function as $A^\pi(s,a) = Q^\pi(s,a) - V^\pi(s)$. Hereafter, we omit the expectation subscripts unless needed for clarity.

\subsection{Policy Gradient and Trust-Region Methods} \label{sec:desc_pg_methods}

In this work, the policy is parameterized by $\theta \in \mathbb{R}^d$, \textit{i.e.} $\pi:=\pi_\theta$ (for instance, $\theta$ can represent a linear function or a neural network), and we focus on a class of RL methods called \textit{policy gradient} methods. Policy gradient methods directly optimize the policy parameters $\theta$ with the objective of maximizing $J(\theta) := J(\pi_\theta)$. The \textit{policy gradient theorem} characterizes the gradients of $J(\theta)$ w.r.t. $\theta$:
\[
    \nabla_\theta J(\theta) = \mathbb{E}\left[Q^\pi(s_t, a_t) \nabla_\theta \log \pi_\theta(a_t \vert s_t)\right] \enspace .
\]
Using sample estimates of the gradients, we can use gradient ascent on the policy parameters to optimize $\pi_\theta$. We typically use estimates of the form
\[
    \hat{g} = \hat{\mathbb{E}}\left[\nabla_\theta\log \pi_\theta(a_t \vert s_t) \hat{A}_t\right] \enspace ,
\]
where $\hat{A}_t$ is an estimator of the advantage function at step $t$. The advantage estimators can be directly computed using Monte Carlo methods (\textit{e.g.}, \textsc{Reinforce} \citep{REINFORCEWilliams92}) or using a trained value function $V_\phi$.

A weakness of policy gradient methods is that small updates in the parameter space may lead to large policy shifts which can harm performance. Natural Policy Gradient (NPG) \citep{NPGKakade} addresses this by rescaling the gradient with the inverse Fisher Information matrix (FIM) $F(\theta) = \mathbb{E}\left[\nabla_\theta \log \pi_\theta(a \vert s) \nabla_\theta \log \pi_\theta(a \vert s)^T \right] \in \mathbb{R}^{d \times d}$, \textit{i.e.}, $\tilde{g} = F(\theta)^{-1} \hat{g}$, taking into account the geometry of the policy space. On the other hand, Trust Region Policy Optimization (TRPO) \citep{TRPOSchulmanLAJM15} uses a hard constraint on the Kullback-Leibler (KL) divergence between the current and the new policy to prevent destructively large policy updates (\textit{i.e.}, a \textit{trust-region} constraint) while optimizing a surrogate loss:
\begin{align*}
    &\max_\theta  L_{\theta_\text{old}}(\theta) := J(\theta) + \mathbb{E}_{s \sim \rho_{\theta_\text{old}}, a\sim \pi_\theta}\left[A^{\theta_{\text{old}}}(s,a) \right] \\&\quad \text{s.t.} \quad \mathbb{E}_{\pi_{\theta_\text{old}}} \left[\text{KL}(\pi_{\theta_\text{old}}, \pi_\theta)\right] \le \alpha \enspace,
\end{align*}

Proximal Policy Optimization (PPO) \citep{PPOSchulmanWDRK17} follows the same logic as TRPO while being significantly easier to implement. It optimizes a surrogate objective which appropriately clips $r_t(\theta) = \frac{\pi_\theta(a_t \vert s_t)}{\pi_{\theta_\text{old}}(a_t \vert s_t)}$, the probability ratio between the current and the new policy:
\[
    L^\text{PPO}(\theta) = \mathbb{E}\left[\min(r_t(\theta)\hat{A}_t, \text{clip}(r_t(\theta), 1 - \epsilon, 1 + \epsilon))\right] \enspace.
\]

\subsection{Differential Privacy} \label{sec:prel_dp}

Machine learning models notoriously leak information about their training data, which can be of great concern in risk-sensitive scenarios. To protect against such leakage, Differential Privacy (DP) \cite{Dwork06} ensures that two models trained on adjacent datasets $D$, $D^\prime$, \textit{i.e.}, differing on only one record (denoted $d(D, D^\prime)$), are probabilistically close. Definition~\ref{def:dp} gives a formal definition of DP.

\begin{definition} \label{def:dp} \emph{$(\epsilon, \delta)$-differential privacy.}
    Given $\epsilon > 0$, $\delta \in [0, 1)$, a \textit{mechanism} $h$ (\textit{i.e.}, a randomized function of the data) is $(\epsilon, \delta)$-DP if for any pair of adjacent datasets $D$, $D^\prime$, and any subset $\mathcal{E}$ in $h$'s range:
    \[
        \mathbb{P} \left(h(D) \in \mathcal{E} \right) \le e^\epsilon \cdot \mathbb{P} \left(h(D^\prime) \in \mathcal{E} \right) + \delta \enspace .
    \]
\end{definition}

To make a given function $f$ private we add noise whose magnitude depends on $f$'s sensitivity $S_f$, that is how much the value of $f$ may change in norm $\Vert \cdot \Vert$ between two adjacent datasets: $S_f = \max_{d(D, D^\prime)=1} \Vert f(D) - f(D^\prime)\Vert$. For $\epsilon, \delta \in (0, 1)$, the Gaussian mechanism adds noise $\zeta \sim \mathcal{N}(0, C(\delta)^2 \cdot S_f/\epsilon^2)$ and is $(\epsilon, \delta)$-DP for $C(\delta) \ge \sqrt{2 \log(1.25/\delta)}$, where $S_f$ is computed w.r.t. the $L_2$-norm.

We can train DP machine learning models by applying the Gaussian mechanism to gradients during optimization, as done by \textsc{DP-SGD} \cite{Abadi_2016} which has achieved great success in some deep learning applications. Since neural network gradients are typically unbounded, \textsc{DP-SGD} clips per-sample gradients $\hat{g}$ with $\bar{g} = \hat{g} / \max(1, \Vert \hat{g} \Vert_2 / S)$, so that $\Vert \bar{g}\Vert_2 \le S$ for $S > 0$. 
%We denote $\bar{S} = 1 / \max(1, \Vert \hat{g} \Vert_2 / S)$ the effective clipping factor. 
Considering \textit{add-or-remove adjacency}, where $D^\prime$ is obtained by adding or removing a record from $D$, the sensitivity of $\bar{g}$ is thus bounded by $S$. For $z > 0$ and $\delta \in (0,1)$, adding noise $\mathcal{N}(0, z^2 S^2 \cdot I_d)$ to $\bar{g}$ ensures that the final model is $(\epsilon, \delta)$-DP for some $\epsilon$ that depends on $z$ and other algorithm parameters. 

% \paragraph{Notations.} For a symmetric positive definite matrix $A = (A_{ij})_{1 \le i,j \le d}$ (denoted $A > 0$), we denote $A^{1/2}$ its unique square root, \textit{i.e.} the unique symmetric matrix $B$ such that $BB=A$. We also denote $\text{tr}(A) = \sum_{i=1}^d A_{ii}$ its trace and $\bar{\sigma}(A)$ its maximum singular value. The operator norm of $A$ is defined as $\Vert A \Vert_\text{op} = \sup_{\Vert v \Vert_2 \le 1}\Vert Av \Vert$, and we have $\Vert Ax \Vert_2 \le \Vert A \Vert_\text{op} \Vert x\Vert_2$ for every $x$. Because $A > 0$, $\Vert A \Vert_\text{op} = \bar{\sigma}(A)$.
\section{Differentially Private On-Policy Policy Gradient with Trust Region} \label{sec:theory_dppg}

\subsection{Differentially Private Personalized RL} \label{sec:dp_perso_rl}

%In this section, we define the threat model for our framework and formulate the definition of \textit{joint differential privacy} (JDP) \citep{DPRLVietriBKW20} in the infinite-horizon setting.

We consider the setting of personalized RL with sensitive data (illustrated in Algorithm~\ref{alg:generic_protocol}), where a system provides personalized recommendations based on sensitive user information. We consider an infinite sequence of users $u=1,2,...$ interacting with the environment. Each user $u$ generates a trajectory $\tau_u = ((s^u_t, a^u_t, r^u_t))_{t \ge 0}$. In the following, we identify a user $u$ with the trajectory $\tau_u$ they generate, and use these terms interchangeably. Both the states $(s_t^u)_{t \ge 0}$ and the rewards $(r_t^u)_{t \ge 0}$ may contain sensitive information about the user. Table~\ref{tab:ex_sensitive_info} in appendix provides different examples. As users' data arrive, the policy is updated with the goal to maximize an objective function $L(\pi)$. 
%When specified, we update the policy using the data collected since the last update (\textit{i.e.}, between two iterations), such that we improve some objective function $L(\pi)$.
Note that users need not arrive sequentially but can interact asynchronously with the environment.

\paragraph{Threat Model.} We denote $D_M = (\tau_u)_{u=1}^M$ the interaction data generated by users $u=1,..., M$. We view the RL training algorithm as a randomized mechanism $\mathcal{T}$ taking interaction data $D_M$ as input and releasing a policy $\pi_M$ as output. The released policy $\pi_M$ can be used by future users $u=M+1,M+2...$ and also be observed by a malicious adversary. Therefore, we do not want $\pi_M$ to reveal sensitive information about users $u=1,...,M$, which can be achieved with differential privacy. 

In the RL literature, the most prevalent notion of DP is \textit{joint differential privacy} (JDP) (Definition~\ref{def:jdp}). JDP states that the data of a user $m$ does not significantly influence the actions recommended to the other users $u \ne m$. Therefore, even if all users, past and future, collude against $m$ by sharing their recommendations (similarly, if an adversary had access to this data), they could not extract more information about $m$'s private data than allowed by the DP bound. 

\begin{definition} \label{def:jdp}
    \emph{(Joint Differential Privacy (JDP)).}
    Given $\epsilon > 0$ and $\delta \in (0,1)$, a mechanism $\mathcal{M}: D \rightarrow \mathcal{A}^{\vert D \vert}$ is ($\epsilon$, $\delta$)-joint DP if for any $M \in \mathbb{N}^\star$, any $u \in [\![1, M]\!]$, any neighboring dataset of interaction data $D_M$, $D_{M}^\prime = D_M \setminus\{\tau_u\}$ differing only on user $u$'s data, and any subset of action sequences $E \subseteq \mathcal{A}^{\vert D_M^\prime \vert}$:
    \[
        \mathbb{P}[\mathcal{M}_{-u}(D_M) \in E] \le e^\epsilon \mathbb{P}[\mathcal{M}_{-u}(D_M^\prime) \in E] + \delta\enspace,
    \]
    where $\mathcal{M}_{-u}(D_M)$ is the sequence of actions recommended to all users $1,...,M$ but $u$.
\end{definition}

While JDP offers strong guarantees by preventing arbitrary collusion, especially in online settings with continuous policy updates, we argue that another definition could be more suitable for certain use cases. Many applications alternate between training phases, where the agent learns from private user data, and exploitation phases, where the policy is deployed for use by other users. Examples include RLHF, health treatment recommendations, and autonomous driving. Thus, a more appropriate privacy objective is to ensure that the released policy (e.g., a fine-tuned LLM in RLHF) does not leak too much information about any specific user data (or trajectory) used during training. To formalize this notion, we propose the definition of \textit{trajectory-level} DP (TDP), stated in Definition~\ref{def:tdp}. TDP explicitly shields the released policy, not merely the actions, therefore protecting against privacy attacks that would exploit information about the policy itself (for instance, \cite{Fu_FineTuningAttack2024} attack fine-tuned LLMs by exploiting the model logits).

\begin{definition} \label{def:tdp}
    \emph{(Trajectory-level Differential Privacy (TDP)).}
    Using the notations from Definition~\ref{def:jdp}, a mechanism $\mathcal{M}: D \rightarrow \Pi$ is ($\epsilon$, $\delta$)-trajectory-level DP if for any subset of policies $E \subseteq \Pi$:
     %Given $\epsilon > 0$ and $\delta \in (0,1)$, a mechanism $\mathcal{M}: D \rightarrow \Pi$ is ($\epsilon$, $\delta$)-trajectory-level DP if for any $M \in \mathbb{N}^\star$, any $u \in [\![1, M]\!]$, any neighboring dataset of interaction data $D_M$, $D_{M}^\prime = D_M \setminus\{\tau_u\}$ differing only on user $u$'s data, and any subset of policies $E \subseteq \Pi$:
     \[
        \mathbb{P}[\mathcal{M}_{-u}(D_M) \in E] \le e^\epsilon \mathbb{P}[\mathcal{M}_{-u}(D_M^\prime) \in E] + \delta\enspace,
    \]
    where $\mathcal{M}(D_M)$ is the policy released and trained on $D_M$.
\end{definition}

Theorem~\ref{thm:generic_tdp_jdp} below states that, if the policy update step $\pi \leftarrow \mathcal{M}(\pi, D^\pi)$ is $(\epsilon, \delta)$-DP at the trajectory level, then Algorithm~\ref{alg:generic_protocol} is both $(\epsilon, \delta)$-TDP and $(\epsilon, \delta)$-JDP. 
Being on-policy, Algorithm~\ref{alg:generic_protocol} only uses the data of a given user for a single policy update before discarding it, which avoids accumulating privacy leakage as a consequence of the sequential composition property of DP. 
By contrast, the privacy guarantees of an off-policy method that stores and reuses data would demand more careful consideration.  
%In this regard, we point out that on-policy methods offer a particular advantage over off-policy methods in a private setting --- where the same data is stored and re-used --- making up for their worse sample efficiency.

%One may notice that Definition~\ref{def:tdp} is different from the notion of \textit{joint differential privacy} (JDP) introduced in \citet{vietri2020private} and subsequently used in the DP RL literature. In fact, it is weaker. Indeed, Definition~\ref{def:tdp} only cares for future outputs that may link information about the $m$-th user, while JDP accounts for both past and future outputs. This means that JDP prevents any arbitrary collusion from users $u \ne m$ against $m$'s data, while in theory TDP could allow past users $u<m$ to submit information to the adversary that would leak information. However, we find Definition~\ref{4.1} to be more intuitive and to make more sense with regard to practical use cases. %RLHF? Personalized services
%Moreover, it appears that making the policy update from Algorithm~\ref{alg:generic_protocol} $(\epsilon, \delta)$-DP also makes Algorithm~\ref{alg:generic_protocol} $(\epsilon, \delta)$-JDP. % Which enables direct comparison with JDP algos from the literature?
%We also want to point out that whatever the definition, it is impossible to ensure that user $u$'s data will be private w.r.t. their own recommendations, since it would defeat the purpose of personalized recommendation. 

\begin{restatable}{theorem}{JDPPersoRL} \label{thm:generic_tdp_jdp}
\emph{($(\epsilon, \delta)$-DP Personalized RL).}
If the update mechanism $\mathcal{M}_\pi:D \longrightarrow\mathcal{M}(\pi, D^\pi) = \pi^\prime$ is $(\epsilon, \delta)$-TDP, then Algorithm~\ref{alg:generic_protocol} is both $(\epsilon, \delta)$-TDP and $(\epsilon, \delta)$-JDP. 
\end{restatable}

\begin{algorithm}[tb]
   \caption{Sensitive Personalized RL (\textit{Training Phase})}
   \label{alg:generic_protocol}
\begin{algorithmic}
   \STATE Initialize $\pi$, iteration $i \leftarrow 0$, on-policy data $D \leftarrow \emptyset$
   \FOR{user $u=1,2,...$}
        \STATE $\tau_u \leftarrow \emptyset$
        \FOR{$t=0,1,2,...$}
            \STATE $u$ sends \textbf{sensitive state} $s_t^u$ to the system $S$
            \STATE $S$ recommends action $a_t^u \sim \pi(\cdot \vert s_t^u)$
            \STATE $S$ and $u$ observe \textbf{sensitive reward} $r_t^u$
            \STATE $\tau_u \leftarrow \tau_u \cup \{(s_t^u, a_t^u, r_t^u)\}$
        \ENDFOR
        \STATE $D^\pi \leftarrow D^\pi \cup \{\tau_u\}$
        \IF{\text{time to update}}
            \STATE Update policy: $\pi \leftarrow \mathcal{M}(\pi, D^\pi)$
            \STATE $i \leftarrow i + 1$, $D^\pi \leftarrow \emptyset$
        \ENDIF
   \ENDFOR
\end{algorithmic}
\end{algorithm}

\subsection{Private Policy Gradient Updates with Trust Region} \label{sec:ppg_updates}

Our goal is now to propose a differentially private policy update $\pi \leftarrow \mathcal{M}(\pi, D^\pi)$. We consider the function approximation setting where the policy is parameterized by $\theta \in \Theta$, \textit{i.e.}, $\pi := \pi_\theta$. As described in Section~\ref{sec:prel_dp}, \textsc{DP-SGD} privatizes the training of machine learning models by injecting noise in the gradients. Therefore, it is sensible to consider a \textit{differentially private policy gradient} (\textsc{DPPG}) algorithm based on the following noisy policy gradient updates:
\[
    \theta = \theta_\text{old} + \eta \cdot \left(\hat{g}(D^{\pi_\theta}) + \xi \right) \enspace,
\]
where $\theta_\text{old}$ are the parameters of the policy at the current iteration (the one used to collect the data), $\hat{g}(D^{\pi_\theta})$ is the sample gradient of $L(\theta) := L(\pi_\theta)$ estimated on on-policy data $D^{\pi_\theta}$, which we denote hereafter $\hat{g}$ for simplicity, and $\xi \sim \mathcal{N}(0, \sigma^2 \cdot I_d)$ is a zero-mean Gaussian noise.
To guarantee differential privacy, the noise scale $\sigma$ must be large enough to hide changes in $\hat{g}$'s magnitude between two neighbor datasets. Unfortunately, without further assumptions on $L(\theta)$, $\hat{g}$ is unbounded. This requires clipping the gradient $\hat{g}$ with some \textit{clipping norm} $S$ (see Section~\ref{sec:desc_pg_methods}), typically a hyperpameter.
%The value of $S$ can be tricky to chose: a too large $S$ will lead to unnecessarily large noise, and a too small $S$ yields too small updates.
Denoting $\bar{S} = 1 / \max(\Vert \hat{g} \Vert / S, 1)$ the effective clipping factor (since we only scale down $\hat{g}$ if its norm is more than $S$), the resulting update $\Delta\theta = \theta - \theta_\text{old}$ is thus:
\begin{equation} \label{eq:dp_pg_update}
    \Delta \theta = \eta \cdot \left(\bar{S}\hat{g} + \xi\right) \enspace.
\end{equation}
Given a noise multiplier $z > 0$, if $\xi \sim \mathcal{N}(0, z^2S^2 I_d)$, the update~\eqref{eq:dp_pg_update} is $(\epsilon, \delta)$-DP w.r.t. data $D^{\pi_\theta}$ for some $\epsilon:=\epsilon(z, \delta)$. 

Equation (\ref{eq:dp_pg_update}) shows that we can use the clipping norm $S$ to control the size of the policy updates. Coincidentally, this is also the underlying idea behind trust-region methods, as described in Section~\ref{sec:desc_pg_methods}: constrain two consecutive policies $\pi_{\theta_\text{old}}$ and $\pi_\theta$ to remain close in some sense. TRPO, for instance, uses the KL divergence as a measure of proximity.
Building on this parallel between trust-region methods and noisy policy gradient updates with clipping, we argue that, for policy gradient methods, the introduction of Differential Privacy can be reduced to the computation of appropriate trust regions (through the clipping norm $S$), thus avoiding the sacrifice of theoretical properties of the DP-less methods, which we develop in the next section.

%In the following, we explore this connection and derive theoretical guarantees regarding the DP policy updates defined in Equation~\ref{alg:dp_pg}. 
In the following, we explore the theoretical properties of DP Policy Gradient algorithms based on the update rule (\ref{eq:dp_pg_update}). We first focus on the standard policy gradient seen as a trust-region method, before delving into TRPO. In particular, we derive insights about how we should set $S$, which plays a crucial role in accommodating the presence of noise in the policy updates.
All proofs and other involved computations from the following sections are relegated in the appendix.

\subsubsection{Private Standard Policy Gradient} \label{sec:spg_theory}

Although not explicitly a trust-region method, the standard policy gradient can be formulated as a trust-region update maximizing a first-order approximation of TRPO's loss $L_{\theta_\text{old}}(\theta)$ using an $L_2$ constraint \cite{TRPOSchulmanLAJM15}:

\begin{align}
    &\max_\theta L^\text{PG} (\theta) := \nabla_\theta L_{\theta_\text{old}}(\theta)^\top(\theta - \theta_\text{old}) \nonumber\\
    \label{eq:stand_pg_constraint}&\quad \text{s.t.} \quad \frac{1}{2}\Vert \theta - \theta_\text{old}\Vert_2^2 \le \alpha \enspace.
\end{align}

In the following, we consider that we have access to the true gradient $g:=\nabla_\theta L_{\theta_\text{old}}(\theta)$. We denote $\bar{g} = \bar{S} g$ the clipped gradient and perform noisy policy gradient updates with 
\[
  \xi \sim \mathcal{N}(0, z^2S^2 I_d) \enspace.
\]

\paragraph{Trust Region constraint with noisy updates.} First, we show that, choosing $S$ appropriately, we are able to ensure that constraint~\eqref{eq:stand_pg_constraint} holds with high probability. We have:
\begin{equation*}
    \frac{1}{2}\Vert \theta - \theta_\text{old}\Vert_2^2 = \frac{\Delta \theta^\top \Delta \theta}{2} = \frac{\eta^2}{2} (\bar{g} + \xi)^\top(\bar{g} + \xi) \enspace.
\end{equation*}
This quadratic form in $\xi$ follows a (scaled) non-central $\chi^2$ distribution.
Proposition~\ref{prop:spg_tr_dist} characterizes this distribution and provides an upper bound on the clipping norm $S$ such that the noisy update remains within the trust region.
\begin{proposition} \label{prop:spg_tr_dist}
    The $L_2$ trust region size follows a scaled non-central $\chi^2$ distribution:
    \begin{equation} \label{eq:clipping_norm_spg_quantile}
        \frac{\Delta \theta^\top\Delta \theta}{2} = \frac{\eta^2 z^2 S^2}{2} \chi^2 \left(d, \nu^2 \right) \enspace,
    \end{equation}
    with $\nu^2 = \Vert \bar{g} \Vert^2 / z^2S^2 \le z^{-2}$. Therefore, given $\alpha \ge 0$ and $\beta \in (0,1)$ and denoting $q_{1-\beta}^{\chi^2(d, \nu^2)}$ the quantile of order $1 - \beta$ of $\chi^2(d, \nu^2)$, if we set:
    \[
        S \le \frac{1}{\eta z} \sqrt{\frac{2 \alpha}{q_{1-\beta}^{\chi^2(d, z^{-2})}}} \enspace,
    \]
    then $\mathbb{P}[\frac{\Delta\theta^\top \Delta\theta}{2} \le \alpha] \ge 1 - \beta$, \textit{i.e.}, the DP policy update remains within the $L_2$ trust region of size $\alpha$ with probability at least $1 - \beta$.
\end{proposition}
Note that the quantiles $q_{1-\beta}^{\chi^2(d, \nu^2)}$ can be easily  computed numerically. Moreover, computing the expectation $\mathbb{E}\left[\frac{\Delta \theta^\top \Delta \theta}{2}\right] =\frac{\eta^2}{2}\left(\Vert \bar{g} \Vert^2 + z^2 S^2 d \right)$ and using the Markov inequality, Proposition~\ref{prop:spg_tr_markov} provides a slightly more interpretable result:
\begin{proposition}\label{prop:spg_tr_markov}
    Given $\alpha \ge 0$ and $\beta \in (0,1)$, if we set:
    \begin{equation}\label{eq:clipping_norm_spg}
        S \le \frac{1}{\eta} \sqrt{\frac{2 \alpha \beta}{1 + z^2d}} \enspace,
    \end{equation}
    then $\mathbb{P}[\frac{\Delta\theta^\top \Delta\theta}{2} \le \alpha] \ge 1 - \beta$.
\end{proposition}

As expected, $S \propto \sqrt{\alpha}$ relates to the size of the trust region, and the larger the probability to remain within the trust region (corresponding to a smaller $\beta$ or a larger $q_{1-\beta}^{\chi^2(d, \nu^2)}$), the smaller we need to set $S$. Stronger DP guarantees (larger $z$) and higher-dimensional problems (larger $d$) also require more conservative updates, which is in line with the intuition. Of course, the value of $S$ is also tied to the learning rate $\eta$, since the latter also determines the size of the update.
% Can we do better?

\paragraph{Objective gap.} We showed that a small enough $S$ keeps the update within the trust region with high probability. %, which is the first guarantee that a DP policy gradient will work.
While this prevents a destructively large update, it tells us nothing about the direction of the update and whether we actually improve the objective. We therefore consider one step of update and denote $\theta^\star = \theta_\text{old} + \eta g$ the parameter obtained with a non-private policy gradient update, and $\tilde{\theta} = \theta_\text{old} + \eta(\bar{g} + \xi)$ the parameter obtained following the DP update. We are interested in the objective gap:

%By construction, the trust region constraint (\ref{eq:stand_pg_constraint}) holds for both $\theta^\star$ and $\tilde{\theta}$.

\[
     L^\text{PG}(\tilde\theta) - L^\text{PG}(\theta^\star) = g^\top (\tilde\theta - \theta^\star) \enspace.
\]

%

% \begin{restatable}{proposition}{SPGLossIneq}
%     For any $\lambda \ge 0$:
%     \[
%         \mathbb{P}\left(L^{PG}(\tilde{\theta}) \le L^{PG}(\tilde{\theta}) - K(\lambda)\right) \le 1 - \frac{\lambda^2}{\eta^2 z^2 S^2 \Vert g\Vert^2 + \lambda^2} ,
%     \]
%     where $K(\lambda) = \lambda + \eta(1 - \bar{S}) \Vert g \Vert^2$.
% \end{restatable}

Proposition~\ref{prop:spg_loss_gap_ineq} shows that, if we set $S$ small enough, we can control the objective gap with high probability. Similarly to previous results, the upper bound is inversely correlated to the learning rate and noise multiplier.

\begin{restatable}{proposition}{SPGLossIneq} \label{prop:spg_loss_gap_ineq}
   Given $\lambda > 0$ and $\beta_2 \in (0, 1)$, if we set:
    \begin{equation} \label{eq:spg_loss_gap_ineq}
        S \le \frac{\lambda}{\eta z \Vert g \Vert}\sqrt{\frac{\beta_2}{1 - \beta_2}} \enspace.
    \end{equation}
    then, denoting $K(\lambda) = (\eta(1 - \bar{S})\Vert g \Vert + \lambda)$:
    \[
        \mathbb{P} \left[L^\text{PG}(\tilde\theta)  \ge L^\text{PG}(\theta^\star) -  K(\lambda)\right] \ge 1 - \beta_2 \enspace.
    \]
\end{restatable}

Therefore, by computing appropriate trust regions, that is by setting $S$ small enough that (\ref{eq:spg_loss_gap_ineq}) and either (\ref{eq:clipping_norm_spg_quantile} or \ref{eq:clipping_norm_spg}) hold, our DP policy gradient preserves the theoretical properties of the non-private policy gradient.

% How to interpret this??

\subsubsection{Private TRPO}

The issue with the previous approach is that the trust region does not take into account the geometry of the policy space: even policy updates that are small within the parameter space can result in large policy shift, potentially hurting performance. TRPO addresses this by formulating a trust region constraint with respect to the (expected) KL divergence between $\pi_{\theta_\text{old}}$ and $\pi_\theta$:
\[
    \mathbb{E}_{\pi_{\theta_\text{old}}} \left[\text{KL}(\pi_{\theta_\text{old}}, \pi_\theta)\right] \le \alpha \enspace.
\]
where the expectation is taken on the states encountered when rolling policy $\pi_{\theta_\text{old}}$. Moreover, in the neighborhood of $\theta_\text{old}$, the KL-divergence can be reasonably approximated using the Fisher information matrix of $\pi_{\theta_\text{old}}$, $F := F(\theta_\text{old})$:
\[
    \mathbb{E} \left[KL(\pi_{\theta_{\text{old}}}, \pi_\theta)\right] \approx \frac{1}{2}(\theta - \theta_\text{old})^\top F (\theta-\theta_\text{old}) := \frac{1}{2}\Delta\theta^\top F\Delta\theta \enspace .
\]
Based on this approximation, we can formulate the trust region constraint in the DP setting, using $\Delta \theta = \eta (\bar{g} + \xi)$,
\begin{equation}\label{eq:fisher_trust_region}
    \frac{\eta^2}{2} (\bar{g} + \xi)^\top F (\bar{g} + \xi) \le \alpha \enspace.
\end{equation}
%(\ref{eq:fisher_trust_region}) directly links the noisy policy gradient update to the trust region constraint of TRPO \textit{via} the FIM. 
% As above, we can extend the left-hand term to analyze its distribution:
% \begin{align*}
%      \eta^2 (\bar{g} + \xi)^\top F (\bar{g} + \xi) = \eta^2 (\bar{g}^\topF\bar{g} + 2\bar{g}^\topF\xi + \xi^\top F \xi).
% \end{align*}

As with standard PG, Proposition~\ref{prop:trpo_kl_tr} characterizes the distribution of (\ref{eq:fisher_trust_region}) and derives an upper bound on $S$ \footnote{If we want to use this bound in practice and need to estimate $F$ or its eigenvalues on private data, then this step should also be privatized. We consider this beyond the scope of the current paper.}:
\begin{proposition} \label{prop:trpo_kl_tr}
    The size of the (approximated) KL trust region follows a generalized non-central $\chi^2$ distribution:
    \[
        \frac{1}{2} \Delta\theta^\top F\Delta \theta = \frac{\eta^2}{2}z^2S^2 \sum_{i=1}^d \sigma_i(F) \chi^2(1, \nu_i^2) \enspace,
    \]
    where $F = P^\top\mathrm{diag}(\sigma_1(F),...,\sigma_d(F))P$ is the spectral decomposition of $F$ and $\nu = (\nu_1,...,\nu_d) = \frac{1}{zS} P^\top\bar{g}$. Moreover, given $\alpha \ge 0$ and $\beta \in (0,1)$, if we set:
    \begin{equation}\label{eq:kl_clipping_norm_bound}
        S \le \frac{1}{\eta} \sqrt{\frac{2\alpha\beta}{\max \sigma_i(F) + z^2\mathrm{tr}(F)}} \enspace ,
    \end{equation}
    then $\mathbb{P}[\Delta\theta^\top F \Delta\theta \le \alpha] \ge 1 - \beta$. %\textit{i.e.}, the DP policy update remains within the KL trust region of size $\alpha$ with probability at least $1 - \beta$.
\end{proposition}
%We point out that if one would need to estimate $F$ or its eigenvalues on private data to use this bound in practice, then this step should also be privatized. 

% An adequate choice of $S$ based on the geometry of the problem thus allows us to control the distribution shift between two consecutive policies. Meanwhile, if the parameter update $\Delta \theta$ is small enough that $L^\text{TRPO}(\theta) \approx L^\text{TRPO}(\theta_\text{old}) + \nabla_\theta L^\text{TRPO}(\theta)^\top \Delta \theta$ (which can result from choosing a small enough trust region based on the inequality $\min \sigma_i(F)\Vert \Delta\theta \Vert^2 \le \Delta\theta^\top F \Delta\theta \le \alpha$), then 
% \[
%      L^\text{TRPO}(\tilde\theta) - L^\text{TRPO}(\theta^\star) \approx g^\top (\tilde\theta - \theta^\star) \enspace ,
% \]
% and we can get similar guarantees as in Section~\ref{sec:spg_theory}.

% \subsubsection{Estimating the Fisher Matrix in the Private setting}

% \textcolor{blue}{Problem: how to estimate the FIM since it can depend on private data? \textbf{Option 1:} privatizing the FIM and/or its eigenvalues. \textbf{Option 2:} obtaining the FIM without using private data. For instance, if we have access to a simulator (trained on public data or privatized). In the following we can assume that we have access to the exact Fisher information.}

\section{Practical Algorithm} \label{sec:practical_dppg}

% In Section~\ref{sec:theory_dppg}, we showed that using a sufficiently small clipping norm $S$ allowed us to control the magnitude of the policy updates, similarly to a trust-region constraint in the DP-less setting. Moreover, when the noise multiplier $z$ is not too large, we showed that the noisy gradient generally aligns with the true gradient, leading toward objective improvement. In other words, the introduction of DP in policy gradient methods with trust region does not need sacrificing their theoretical properties.
% Although theoretically grounded, the approaches derived in the last section either do not take into account the geometry of the policy space or require the estimation of the FIM, which can be difficult in high-dimensional problems and also poses privacy concerns.
% Thus, we will henceforth treat the clipping norm $S$ as a hyperparameter, understanding that a sufficiently small value preserves the desired theoretical properties. We then propose a practical DP PG algorithm that avoids involved clipping norm computations at each step. While this approach diverges from the theory, it remains efficient and is significantly easier to implement in practice.

%\subsection{\textsc{DP-PG}: Differentially Private Policy Gradient}

After studying the theoretical properties of differentially private policy gradient methods in Section~\ref{sec:theory_dppg}, we now introduce a practical approach based on the generic personalized RL protocol described in Section~\ref{sec:dp_perso_rl}, described in Algorithm~\ref{alg:dp_pg}.

\begin{algorithm}[tb]
   \caption{DP Policy Gradient}
   \label{alg:dp_pg}
\begin{algorithmic}
   %\STATE {\bfseries Input:} noise multiplier $z$, clipping norm $S$, global update frequency $K$, parameters for local update computation $\zeta$, algorithm $\mathcal{A}$ (\textit{e.g.}, PPO)
   \STATE Initialize $\theta$, iteration $i=0$, active users $\mathcal{U}_i \leftarrow \emptyset$
   \FOR{user $u=1,2,...$}
        \STATE $\mathcal{U}_i \leftarrow \mathcal{U}_i \cup \{u\}$
        \STATE Run policy $\pi_{\theta}$ in environment for $T$ timesteps and collect trajectory $\tau_u = ((s^u_t, a^u_t, r_t^u))_{t=1}^T$
        \STATE Compute advantage estimates $\hat{A}_u = (\hat{A}^u_t)_{t=1}^T$
        \STATE Compute clipped update $\bar{g}_u$ from $u$'s data: $g_u = \text{\textsc{ComputeLocalUpdate}}(\theta, \tau_u, \hat{A}_u, S)$
        \IF{u \% K = 0}
            \STATE $i \leftarrow i + 1$
            \STATE $\bar{g} \leftarrow \sum_{u \in \mathcal{U}_i} \bar{g}_u/ K$
            \STATE $\tilde{g} \leftarrow \bar{g} + \mathcal{N}\left(0_d, (\frac{zS}{K})^2I_d\right)$
            \STATE $\theta \leftarrow \theta + \eta \tilde{g}$
            \STATE $\mathcal{U}_i \leftarrow \emptyset$
        \ENDIF
   \ENDFOR
\end{algorithmic}
\end{algorithm}

In Algorithm~\ref{alg:dp_pg}, the policy is only updated every $K$ users (we refer to the steps between two policy updates as an iteration). Therefore, all users from iteration $i$, \textit{i.e.}, users $\mathcal{U}_i = [\![i \cdot K, (i + 1) \cdot K]\!]$), interact with the environment following the same policy $\pi_{\theta_i}$. 
To guarantee privacy at the user level, we compute for each $u \in \mathcal{U}_i$ a local update $\bar{g}_u$ based on $u$'s data only, which is then clipped so that $\Vert \bar{g}_u \Vert \le S$, adequately limiting the influence of $u$ in the global policy update. We treat $S$ as a tunable hyperparameter, hence avoiding the computation of the upper bounds derived in Section~\ref{sec:theory_dppg}, as they either do not take into account the geometry of the policy space or require the estimation of the FIM, which can be difficult in high-dimensional problems and also poses privacy concerns. The clipped updates $\{\bar{g}_u\}_{u \in \mathcal{U}_i}$ are then aggregated as $\bar{g} = K^{-1} \sum_{u \in \mathcal{U}_i} \bar{g}_u$, which has sensitivity $\mathbf{S} = S / K$. We then apply the Gaussian mechanism with noise magnitude $\sigma = z \cdot \mathbf{S}$ to $\bar{g}$ before updating the policy. 

$\bar{g}_u$ is computed through the auxiliary function \textsc{ComputeLocalUpdate}, which depends on the underlying policy gradient methods. Our algorithm is flexible enough that it just requires \textsc{ComputeLocalUpdate} to output an estimate of the ascent direction with norm less than $S$. Algorithm~\ref{alg:local_grad_ppo} in appendix describes the local update function for PPO. Since we compute the final update applied to the policy $\pi_\theta$ based on the aggregation of the clipped local updates $\{g_u\}_{u \in \mathcal{U}_i}$, we can leverage several rounds of updates based on a single user's data within the \textsc{ComputeLocalUpdate} function without incurring more privacy cost, on the same model as \cite{McMahanRT018}.

\paragraph{Privacy guarantees.} It is straightforward to show that the one-step policy update of Algorithm~\ref{alg:dp_pg} is $(\epsilon, \delta)$-TDP, where $\epsilon := \epsilon(z, \delta)$, as a direct application of the Gaussian mechanism at the trajectory level. Following Theorem~\ref{thm:generic_tdp_jdp}, we can derive the privacy guarantees of the full algorithm. Given $\delta \in (0,1)$ and a noise multiplier $z > 0$ (a tunable hyperparameter of Algorithm~\ref{alg:dp_pg}), we then want to compute the privacy budget $\epsilon$. Many works in the literature overlook the fact that the guarantees of the standard Gaussian mechanism from Section~\ref{sec:prel_dp} only hold for $\epsilon \in (0,1)$. However, for $z \ge C_1(\delta) = \sqrt{2 \ln(1.25/\delta)}$ (that is a theoretical $\epsilon \ge 1$), the privacy guarantees do not hold. Fortunately, \cite{improved_gaussian_mechanisms} provide improved Gaussian mechanisms for $\epsilon \ge 1$, which we use to compute the privacy budget of Algorithm~\ref{alg:dp_pg} for arbitrarily large $z$. Section~\ref{sec:app_priv_guarantees_dp_pg} in appendix details the computation of the privacy guarantees for Algorithm~\ref{alg:dp_pg}. In particular, Figure~\ref{fig:z_eps_relation} plots the relationship between $z$ and the total privacy budget $\epsilon$ spent by Algorithm~\ref{alg:dp_pg}.

%In the following, we further make several remarks regarding Algorithm~\ref{alg:dp_pg}.

\begin{remark} \textbf{Delayed updates.}
    Using delayed updates allows us to decrease the amount of noise added by a factor $K$. Algorithm~\ref{alg:dp_pg} assumes a constant policy update frequency $K$ for simplicity, but this needs not be the case. Moreover, when simulators are available, PG methods often use several parallel environments to speed up training and use larger batches. Using delayed updates in our user-oriented setting is consistent with this feature of PG methods.
\end{remark}

\begin{remark} \textbf{DP clipping and \textsc{PPO} clipping.}
    \textsc{PPO} clips the ratio $r_t(\theta) = \frac{\pi_\theta(a_t\vert s_t)}{\pi_{\theta_\text{old}}(a_t\vert s_t)}$. We found that when clipping the gradients for DP purposes, further clipping $r_t(\theta)$ was not necessary. This supports our theoretical findings suggesting that clipping the gradient was akin to the computation of trust regions, and avoid using too many hyperparameters.
\end{remark}

%We find that it is easier to use $z$ as a hyperparameter and computing $\epsilon$ \textit{a posteriori}.

\section{Experiments} \label{sec:exps}

We now evaluate our DP Policy Gradient algorithm from Section~\ref{sec:practical_dppg}, referred to as \textsc{DPPG}, on several benchmarks: \textsc{Riverswim} \cite{Riverswim_OsbandRR13}, Gym Control's \textsc{CartPole} and \textsc{Acrobot} \cite{openai_gym}, MuJoCo's \textsc{HalfCheetah} and \textsc{Hopper} \cite{mujoco}, as well as a personalized medication dosing simulator for diabetic patients, referred to as \textsc{Dosing} \cite{dosing}.
We also apply our approach to an RLHF sentiment tuning task.
More details regarding experiments are provided in appendix (Section~\ref{sec:app_exps_details}). All average values and confidence intervals are computed on at least 10 random seeds. Table~\ref{tab:asymptotic_results} and Figure~\ref{fig:all_plots} present asymptotic results and learning curves for all environments (except \textsc{Riverswim}).

\paragraph{Riverswim.}

\begin{figure}[t]
\centering
\begin{minipage}{.49\textwidth}
%\vspace{0pt}
\includegraphics[width=\linewidth]{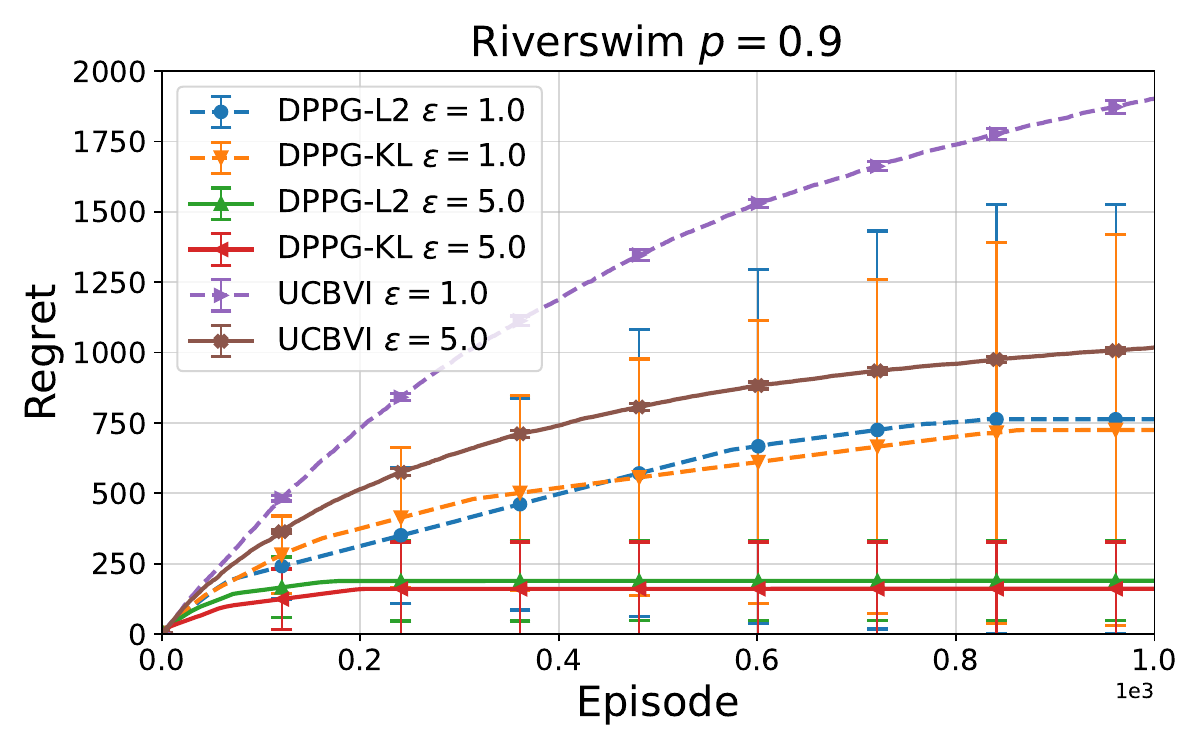}
\end{minipage}
\caption{Cumulative regret on \textsc{Riverswim} for $\epsilon=1.0$ (\textit{dashed line}) and $\epsilon=5.0$ (\textit{solid line}).}
\label{fig:riverswim}
\end{figure}

\begin{figure*}[t]
\centering
\begin{minipage}{.49\textwidth}
%\vspace{0pt}
\includegraphics[width=\linewidth]{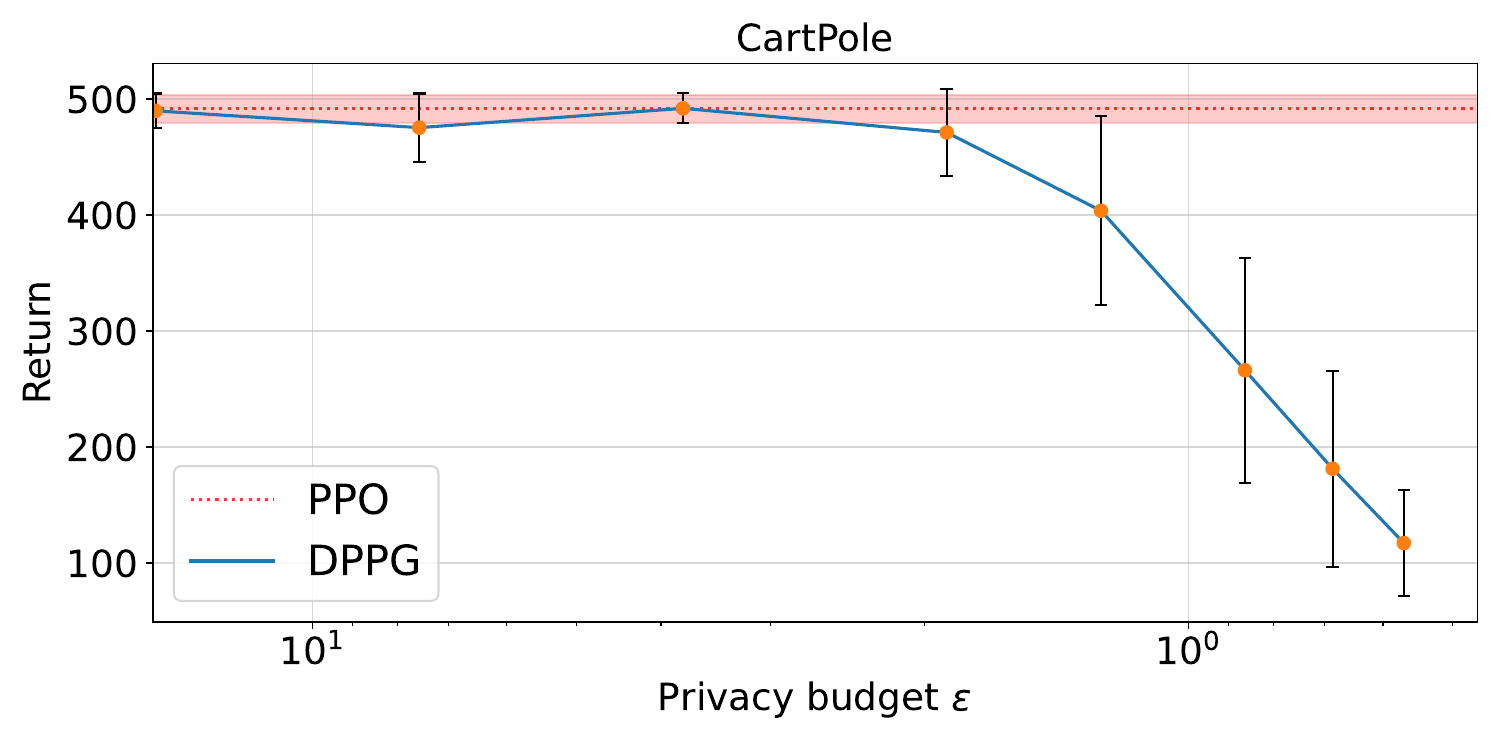}
\end{minipage}
\begin{minipage}{.49\textwidth}
%\vspace{0pt}
\includegraphics[width=\linewidth]{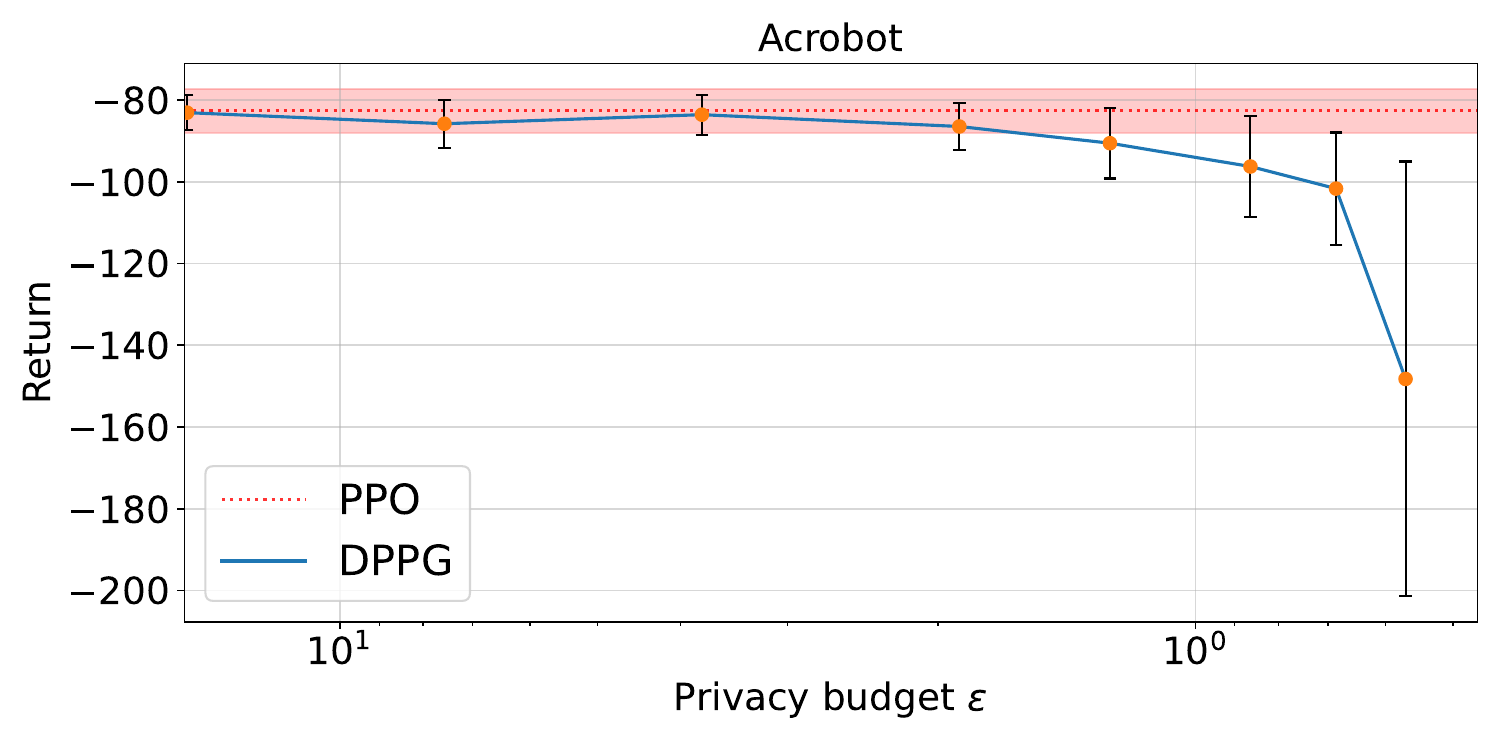}
\end{minipage}
%\hfill
%\vspace{-2pt}
\caption{Asymptotic performance vs. privacy budget $\epsilon$ \textsc{CartPole} (\textit{left}) and \textsc{Acrobot} (\textit{right}) in log scale.}
\label{fig:control_f_eps}
\end{figure*}
\vspace{-.5cm}
\begin{table*}[t]
\scriptsize
\vskip 0.15in
\begin{center}
\begin{sc}
\begin{tabular}{c@{\hskip 20pt}cc@{\hskip 20pt}lc@{\hskip 10pt}ll}
\toprule
Env. & \textsc{PPO} & \textsc{A2C} & \multicolumn{4}{c}{\textsc{DP Policy Gradient}} \\
\midrule
\textsc{CartPole} & $493.2 \pm 9.4$ & - & $\mathbf{z=1.0}$ & $496.4 \pm 6.6$ & $\mathbf{z=3.0}$ & $375.5 \pm 86.7$ \\
\textsc{Acrobot} & $-82.1 \pm 5.1$ & - & $\mathbf{z=1.0}$ & $-83.0 \pm 4.4$ & $\mathbf{z=3.0}$ & $-89.8 \pm 7.6$ \\
%\textsc{Inverted-Pendulum} & $897.6 \pm 82.5$ &  & $\mathbf{z=0.05}$ & $288.9 \pm 250.5$ & $\mathbf{z=0.1}$ & $551.6 \pm 309.9$ \\
\textsc{HalfCheetah} & $2105.5 \pm 627.9$ & $1003.3 \pm 27.2$ & $\mathbf{z=0.05}$ & $1548.9 \pm 399.1$ & $\mathbf{z=0.1}$ & $1328.4 \pm 306.5$ \\
\textsc{Hopper} & $2208.2 \pm 477.6$ & $625.0 \pm 106.8$ & $\mathbf{z=0.05}$ & $645.0 \pm 176.4$ & $\mathbf{z=0.1}$ & $808.9 \pm 321.65$ \\
\textsc{Dosing-0} & $78.9 \pm 47.7$ & - & $\mathbf{z=0.1}$ & $109.9 \pm 4.8$ & $\mathbf{z=0.5}$ & $17.2 \pm 57.6$ \\
\textsc{Dosing-1} & $99.5 \pm 51.9$ & - & $\mathbf{z=0.1}$ & $20.7 \pm 53.4$ & $\mathbf{z=0.5}$ & $-24.9 \pm 58.8$ \\
%\textsc{RLHF} & 2.3 & - & $\mathbf{z=0.01}$ & 1.6 & - \\
\bottomrule
\end{tabular}
\end{sc}
\end{center}
\vspace{-2pt}
\caption{Asymptotic performance of \textsc{DPPG} along with non-private baselines.}
\label{tab:asymptotic_results}
\vspace{-10pt}
\end{table*}

We first tackle the tabular MDP \textsc{Riverswim}. It is a standard environment in DP RL where finding an optimal solution requires a good exploration. On this task, we assess \textsc{DPPG} against \textsc{UCB-VI} \cite{chowdhury_differentially_2021}. Results are presented in Figure~\ref{fig:riverswim} and \ref{fig:riverswim_lin}. We use a simple version of \textsc{DPPG} where the policy is log-linear, and where we update the policy after each episode ($K=1$) with a single DP policy gradient ascent step as Equation~\ref{eq:dp_pg_update}. Moreover, we use the theoretical values derived in Section~\ref{sec:theory_dppg} to compute the clipping norm $S$ at each update step. We use two versions of the environment corresponding to different probabilities $p \in \{0.6, 0.9\}$ of obtaining the maximal reward when taking the optimal action at the rightmost state, which influences the difficulty of the task. Moreover, we consider two privacy budgets $\epsilon \in \{1.0, 5.0\}$ and two versions of our algorithm: \textsc{DPPG-KL}, where $S$ is computed according to the KL trust region (Proposition~\ref{prop:trpo_kl_tr}), and \textsc{DPPG-L2} for which $S$ is computed according to an $L_2$ trust region (Proposition~\ref{prop:spg_tr_dist}). For \textsc{DPPG-KL}, we consider that we can approximate $F$ on public data.
Despite greater instability, we observe that, on average, both versions of our algorithm far outperforms \textsc{UCB-VI}. \textsc{DPPG-KL} also performs slightly better than \textsc{DPPG-L2}, suggesting that considering the policy space geometry is beneficial.

\paragraph{Control tasks.} For Gym and MuJoCo control tasks, we use 2-layer neural network policies and deploy the \textsc{DPPG} algorithm with PPO updates as described in Section~\ref{sec:practical_dppg}. We compare our approach to non-private \textsc{PPO}, expecting a drop in performance as we strengthen the privacy guarantees. To quantify the privacy-utility trade-off for \textsc{CartPole} and \textsc{Acrobot}, we evaluate the achieved return for different privacy budgets $\epsilon$ (corresponding to various noise multipliers $z$). 
Results are reported in Figure~\ref{fig:control_f_eps}.% where the average return obtained with non-private PPO (along with the corresponding confidence interval) is represented with a red horizontal line.
We observe that we achieve privacy at almost no cost until $\epsilon$ approaches 1. The cost becomes more significant for $\epsilon \ll 1$, especially on \textsc{CartPole}. Despite offering weak theoretical guarantees, we point out that, given the restrictive worst-case nature of DP, achieving $1\le \epsilon \lessapprox 10$ is a realistic and widely accepted practical privacy goal in the DP ML literature (as documented in \citet{ponomareva_how_2023}).
Achieving privacy in the higher-dimensional continuous control tasks \textsc{Hopper} and \textsc{HalfCheetah} proves to be more costly, and we therefore use weaker privacy levels for these environments. Nonetheless, our approach is still able to retain most of the performance of non-private PPO, especially on \textsc{HalfCheetah}.
%For these tasks, we also compare with \textsc{A2C}. 
Interestingly, \textsc{DPPG} even outperforms \textsc{A2C}, although the latter offers no privacy guarantee. Overall, these results show that our approach is able to retain good performance while achieving relevant DP levels on standard control tasks with NN policies, which, to the best of our knowledge, has not been reported in the DP RL literature before.

\paragraph{Medication Dosing Simulations.} To better illustrate the real-world relevance of our approach, we learn DP personalized medication policies trained on simulated patient data. In the \textsc{Dosing} task, the goal is to keep blood glucose at tolerable levels in Type-1 diabetic patients, and the RL agent controls the insulin uptake. Each trajectory corresponds to a full day of patient measures, and the state contains various health measures such as meal disturbance amounts, insulin levels and blood glucose levels, which can be considered as sensitive and private information. 
We want to protect this information at the patient (or trajectory) level, which is achieved by our TDP policy gradient approach.
We use two versions of the environment: \textsc{Dosing-0} and \textsc{Dosing-1}, described in appendix.
%We use two versions of the environment: \textsc{Dosing-0}, where the meal pattern is constant across episodes, and the more difficult \textsc{Dosing-1} where each patient (or episode) has a different meal pattern throughout the day. 
We use a 2-layer neural network policy with continuous action under two different privacy levels: $z=0.1$ and $z=0.5$ corresponding to $\epsilon \approx 38$ and $\epsilon \approx 8$, respectively. While these tasks are more complex than the previous ones (higher-dimensional state and continuous action space), results show that our approach achieves these privacy levels with limited impact on performance.

\paragraph{RLHF.} Motivated by recent breakthroughs in the field of LLMs where RL played a key role, we also demonstrate the potential of our approach for private LLMs fine-tuning with RLHF. Our task consists of fine-tuning a \textsc{GPT-2} model on a dataset containing IMDB reviews. The goal is to encourage the model to produce positive review continuations, using a BERT sentiment classifier as a reward model. We compare \textsc{PPO} with \textsc{DPPG} with $z=10^{-2}$ as the reinforcement algorithm, and observe on Figure \ref{fig:gpt2} that our approach is still able to achieve large rewards despite the privacy constraints.

\section{Discussion}

%In this work, we demonstrated that introducing differential privacy into policy gradient methods comes at little cost, unlocking the potential of these methods for practical DP RL. Our approach indeed overcomes structural limitations of existing methods, applying to general MDPs and supporting neural networks, hence moving beyond the predominantly theoretical focus of existing work. Empirically, our method achieves strong privacy-utility trade-offs on environments previously unexplored in the literature and shows promising capabilities on complex problems. This work lays the foundation for tackling higher-dimensional tasks --- critical for addressing pressing privacy threats in RL, for instance in LLMs. Beyond scaling up these methods, we believe future work should also focus on exploring comparable off-policy methods and developing practical privacy attacks to correctly assess the guarantees of future DP RL algorithms.

While private RL has so far proved challenging, limiting the literature to mainly theoretical work, we showed that the introduction of differential privacy in policy gradient methods comes at a limited cost, opening encouraging perspectives for the much-needed deployment of private RL agents in real work. Indeed, we proposed an approach that is free from the structural limitations of existing methods, applicable to general MDPs, and compatible with neural network approximations, and showed that it can achieve good privacy-utility tradeoffs on a wide range of tasks not yet addressed in the literature. Opening new perspectives in practical private RL, we believe that this work should be a starting point for addressing higher-dimensional problems, such as the most pressing problem of privacy in LLMs. Moreover, it will be crucial to study practical privacy attacks in deep RL in order to evaluate the privacy capabilities of future algorithms. We also think that future work should be interested in developing similar off-policy methods for use cases where it is most appropriate.

%%%%%%%%%%%%%%%%%%%%%%%%%%%%%%%%

\newpage

% Acknowledgements should only appear in the accepted version.
%\section*{Acknowledgements}

\section*{Impact Statement}

Although RL is increasingly used in real-world applications, research has paid relatively little attention to practical solutions for ensuring data privacy, despite growing evidence that policies trained on personal data can leak sensitive information. As a result, existing solutions are nowhere near effectively addressing privacy concerns in real-world scenarios. This work aims to provide a scalable approach to differentially private RL, overcoming the limitations of existing methods and paving the way for the crucial deployment of private RL agents in practice.

% As recent advances in the field have moved reinforcement learning closer to widespread real-world application, from healthcare to autonomous driving, and as many works have shown that it is no more immune to privacy attacks than any other area in machine learning, it has become crucial to design algorithmic techniques that protect user privacy. In this paper, we contribute to this endeavor by introducing a new approach to privacy in offline RL, tackling more complex control problems and thus paving the way towards real-world private reinforcement learning. We firmly believe in the importance of pushing the boundaries of this research field and are hopeful that this work will contribute to practical advancements in achieving trustworthy machine learning.

\bibliography{main}
\bibliographystyle{icml2025}

%%%%%%%%%%%%%%%%%%%%%%%%%%%%%%%%%%%%%%%%%%%%%%%%%%%%%%%%%%%%%%%%%%%%%%%%%%%%%%%
%%%%%%%%%%%%%%%%%%%%%%%%%%%%%%%%%%%%%%%%%%%%%%%%%%%%%%%%%%%%%%%%%%%%%%%%%%%%%%%
% APPENDIX
%%%%%%%%%%%%%%%%%%%%%%%%%%%%%%%%%%%%%%%%%%%%%%%%%%%%%%%%%%%%%%%%%%%%%%%%%%%%%%%
%%%%%%%%%%%%%%%%%%%%%%%%%%%%%%%%%%%%%%%%%%%%%%%%%%%%%%%%%%%%%%%%%%%%%%%%%%%%%%%
\newpage
\appendix
\onecolumn

\section{Proofs}

\JDPPersoRL*

\begin{proof} We first prove that Algorithm~\ref{alg:generic_protocol} is TDP, before proving it is also JDP.%, whose definition we recall below:

%     \begin{definition} \label{def:jdp}
%     \emph{(Joint Differential Privacy (JDP)).}
%     Given $\epsilon > 0$ and $\delta \in (0,1)$, a mechanism $\mathcal{M}: D \rightarrow \mathcal{A}^{\vert D \vert}$ is ($\epsilon$, $\delta$)-joint DP if for any $M \in \mathbb{N}^\star$, any $u \in [\![1, M]\!]$, any neighboring dataset of interaction data $D_M$, $D_{M}^\prime = D_M \setminus\{\tau_u\}$ differing only on user $u$'s data, and any subset of action sequences $E \subseteq \mathcal{A}^{\vert D_M^\prime \vert}$:
%     \[
%         \mathbb{P}[\mathcal{M}_{-u}(D_M) \in E] \le e^\epsilon \mathbb{P}[\mathcal{M}_{-u}(D_M^\prime) \in E] + \delta\enspace,
%     \]
%     where $\mathcal{M}_{-u}(D_M)$ is the sequence of actions recommended to all users $1,...,M$ but $u$.
% \end{definition}

    \textbf{Algorithm~\ref{alg:generic_protocol} is TDP.}

    For any user $u$, denoting $i_u$ the corresponding iteration, $u$'s data is only used in the $i_u$-th update step before being discarded. We denote $D_i$ the data collected at iteration $i$. In essence, we can view Algorithm~\ref{alg:generic_protocol} as a repetitive application of an $(\epsilon, \delta)$-TDP mechanism on the disjoint data $D = \cup_{i \ge 0} D_i$, where each $u$'s appears in a single $D_i$. By the parallel composition property of DP, Algorithm~\ref{alg:generic_protocol} is therefore $(\epsilon, \delta)$-TDP.
    
    More formally, at iteration $i=0$, we start from a randomly initialized policy $\pi_0$, which is private by definition. Since the mechanism $\mathcal{M}$ is $(\epsilon, \delta)$-TDP, the update $\pi_1 = \mathcal{M}(\pi_0, D_0)$ is also $(\epsilon, \delta)$-TDP. Users $\mathcal{U}_0$'s data is discarded at the end of iteration $0$ and does not appear in subsequent iterations. Moreover, thanks to the \textit{post-processing} property of DP, no application can deteriorate the privacy guarantees of $\pi_1$. Therefore, at iteration $i=1$, $\mathcal{M}$ is applied to a private quantity computed with $(\epsilon, \delta)$-TDP ($\pi_1$) and new user data $D_1$ that is disjoint from $D_0$, that is $D_1 \cap D_0 = \emptyset$. Therefore, by the \textit{parallel composition} property of DP, the update $\pi_2 = \mathcal{M}(\pi_1, D_1)$ is $(\epsilon, \delta)$-TDP. By induction, Algorithm~\ref{alg:generic_protocol} is $(\epsilon, \delta)$-TDP.

    Moreover, we point out that the above analysis does not depend on the number of iterations, and therefore at each iteration $i$, the policy $\pi$ has been computed with an $(\epsilon, \delta)$-TDP algorithm. Every policy released during Algorithm~\ref{alg:generic_protocol} is therefore $(\epsilon, \delta)$-TDP.

    \textbf{Algorithm~\ref{alg:generic_protocol} is JDP.}

    For this proof, we rely on the \textit{billboard lemma} \cite{HsuHRRW16}, formalized in RL in Lemma 2 from \citet{vietri2020private}. Informally, it states that an algorithm is JDP if, for a user $u$, the output sent to user $u$, \textit{i.e.}, the actions $(a^{(u)}_t)_t$ recommended to $u$ by the current policy, is a function of 1) the $u$'s private data, \textit{i.e.}, $(s_t^{(u)}, r_t^{(u)})_t$, and 2) a common signal computed with standard differential privacy, \textit{i.e.}, the private policy $\pi_{i_u}$ at iteration $i_u$. 
    
    Trajectory-level DP (TDP) is no more than standard DP applied at level of trajectories, and we just showed above that if the update mechanism $\mathcal{M}$ is $(\epsilon, \delta)$, the released policy at each iteration is $(\epsilon, \delta)$-TDP. Based on the hypothesis that a user $u$ keeps its own data private, Algorithm~\ref{alg:generic_protocol} is therefore $(\epsilon, \delta)$-JDP from the \textit{billboard lemma}.

    \textbf{Algorithm~\ref{alg:generic_protocol} is JDP prevents against arbitrary collusion.}
    
    One property of JDP is to prevent arbitrary collusion against a given user $m$, that is, if all users $u \ne m$ gather their private data, JDP ensures they cannot recover more information about user $m$ than what is implied by the theoretical privacy guarantees (Section 2.2 from \cite{vietri2020private} provides an example of how users may collude in an online setting). Since Algorithm~\ref{alg:generic_protocol} is JDP, it directly follows that it prevents collusions from users from past and previous iterations: that is, if we denote $\mathcal{U}_i$ the users from iteration $i$, and $\mathcal{U}_{<i}$, $\mathcal{U}_{>i}$ the users from past and future iterations, respectively, users $\mathcal{U}_{<i} \cup \mathcal{U}_{>i}$ cannot colide against users $\mathcal{U}_i$. However, it may be less clear that it prevents collusion of users from a single iteration, which we thus demonstrate below.

    At iteration $i$, let us assume that users $\mathcal{U}_i\setminus\{m\}$ wants to collide against user $m$. This means that users $\mathcal{U}_i\setminus\{m\}$ gather their data $D_{-m} = \cup_{u \in \mathcal{U}_i\setminus\{m\}} D_u$, where $D_u$ is user $u$'s data. Now, let us denote $\mathcal{M}_\text{non-priv}$ the non-private policy update mechanism, and $f$ the privatization mechanism such that $\mathcal{M} = f \circ \mathcal{M}_\text{non-priv}$ is the private update mechanism.

    The private update that is released and that contains user $m$'s private data is: 
    \[
        \mathcal{M}(\cup_{u \in \mathcal{U}_i} D_u) = \mathcal{M}(D_m \cup D_{-m}) = f(\mathcal{M}_\text{non-priv}(D_m \cup D_{-m})) \enspace.
    \]
    On the other hand, gathering their data, users $\mathcal{U}_i\setminus\{m\}$ can compute the corresponding non-private update $\mathcal{M}_\text{non-priv}(D_{-m})$. To recover information about $m$'s data from $ \mathcal{M}(\cup_{u \in \mathcal{U}_i} D_u)$ using $\mathcal{M}_\text{non-priv}(D_{-m})$, they would need to compose the two latter quantities with some function $g(\cdot, \cdot)$:
    \[
        g\left(f(\mathcal{M}_\text{non-priv}(D_m \cup D_{-m})), \mathcal{M}_\text{non-priv}(D_{-m})\right) \enspace.
    \]
    However, because $\mathcal{M}(\cup_{u \in \mathcal{U}_i} D_u) = f(\mathcal{M}_\text{non-priv}(D_m \cup D_{-m}))$ is DP and because no composition with a function $g$ can deteriorate the DP guarantees (\textit{i.e.}, \textit{post-processing}), such collusion protocol cannot break the DP guarantees of the private policy update; Therefore, Algorithm~\ref{alg:generic_protocol} with $(\epsilon, \delta)$-TDP update mechanism also prevents against arbitrary collusion from users of the same iteration.

    \textit{Example with private policy gradient update.} Let us illustrate this with the private policy gradient update from Algorithm~\ref{alg:dp_pg}:
    \[
        \theta \leftarrow \theta + \eta (\bar{g} + \xi), \quad \xi \sim \mathcal{N}(0, z^2S^2/K) \enspace.
    \]
    In this case, the released private quantity containing information about user $m$ is $\tilde{g} = \bar{g} + \xi = \frac{1}{K}\sum_{u \in \mathcal{U}_i} \bar{g}_u + \xi$. Gathering their information, users $\mathcal{U}_i \setminus \{m\}$ can compute $\frac{1}{K} \sum_{u \in \mathcal{U}_i \setminus \{m\}} \bar{g}_u$ and try to infer the gradient $\bar{g}_m$ (which would reveal private information about user $m$) with:
    \[
        \tilde{g} - \frac{1}{K} \sum_{u \in \mathcal{U}_i \setminus \{m\}} \bar{g}_u = \bar{g}_m / K + \xi \enspace.
    \]
    But since $\Vert \bar{g}_m \Vert_2 \le S$, the sensitivity of $\bar{g}_m / K$ is less than $S/K$ in the \textit{add-or-remove} privacy setting. Because $\xi$ is calibrated with sensitivity $S/K$, $\bar{g}_m / K + \xi$ is therefore as private as $\tilde{g}$. Therefore, users $\mathcal{U}_i \setminus \{m\}$ cannot recover more information by colluding that they can get directly from the released quantity $\tilde{g}$.
    
\end{proof}

%%%%%%%%%%%%%%%%%%%%%%%%%

\SPGLossIneq*

\begin{proof}
    We are interested in the following quantity:
\[
     L^\text{PG}(\tilde\theta) - L^\text{PG}(\theta^\star) = g^T (\tilde\theta - \theta^\star) \enspace.
\]

Since $\tilde\theta - \theta^\star = \eta(\bar{g} + \xi) - \eta g = \eta (\bar{S}g + \xi) - \eta g = \eta[(\bar{S}-1)g + \xi]$, we have:
\begin{align*}
    L^\text{PG}(\tilde\theta) - L^\text{PG}(\theta^\star) &= \eta \cdot g^T[(\bar{S}-1)g + \xi] \\
    &= \eta g^T \xi + \eta (\bar{S} - 1) \Vert g \Vert^2 \enspace,
\end{align*}
with $\eta g^T\xi \sim \mathcal{N}(0, \sigma^2)$.

Therefore:
\[
    L^\text{PG}(\theta^\star) - L^\text{PG}(\tilde\theta) = - \eta g^T\xi + \eta(1 - \bar{S})\Vert g\Vert^2 \enspace.
\]

By symmetry, $- \eta g^T\xi$ has the same distribution as $\eta g^T\xi$. Therefore, denoting $\sigma^2 = \eta^2 z^2 S^2 \Vert g \Vert^2$ and applying Cantelli's inequality for $\lambda > 0$:
\begin{align*}
    \mathbb{P}\left(L^\text{PG}(\theta^\star) - L^\text{PG}(\tilde\theta)  - \eta(1 - \bar{S})\Vert g \Vert^2 \ge \lambda\right) &= \mathbb{P}\left(L^\text{PG}(\tilde\theta)  \le L^\text{PG}(\theta^\star) -  (\eta(1 - \bar{S})\Vert g + \lambda)\right) \\
    &= \mathbb{P}\left(L^\text{PG}(\tilde\theta)  \le L^\text{PG}(\theta^\star) -  K(\lambda)\right) \\
    &\le \frac{\sigma^2}{\sigma^2 + \lambda^2} \enspace.
\end{align*}

We want $S$ such that $\sigma^2 / (\sigma^2 + \lambda^2) \le \beta_2 \Longleftrightarrow \sigma^2 \le \lambda^2 \cdot \beta_2/(1-\beta_2)$, which gives us:
\[
    S \le \frac{\lambda}{\eta z \Vert g \Vert}\sqrt{\frac{\beta_2}{1 - \beta_2}} \enspace.
\]

\end{proof}

%%%%%%%%%%%%%%%%%%%%%%%%%

\subsection{Trust Region Constraints (Section~\ref{sec:ppg_updates})} 

In this section, we derive the results regarding the random trust region constraints as analyzed in Section~\ref{sec:ppg_updates}. We first derive the first two moments of the trust region size in the two cases considered (standard policy gradient and TRPO), before deriving the full distributions.

\subsubsection{Expectation and Variance}

We focus on the KL divergence constraint $\frac{1}{2}\Delta \theta^T F \Delta \theta \le \alpha$ as the results for the L2 constraint $\frac{1}{2}\Delta\theta^T\Delta \theta$ can be derived directly from considering taking $I_d/2$ instead of $F$.

We consider the noisy gradient $\tilde{g} = \bar{g} + \xi = \bar{g} + zS \cdot\zeta$, where $\zeta \sim \mathcal{N}\left(0, I_d\right)$. $\bar{g}$ is clipped with constant $S$, \textit{i.e.}, $\Vert \bar{g} \Vert_2 \le S$. We have:
\begin{align}
    \frac{\eta^2}{2} \tilde{g} F \tilde{g} &= \frac{\eta^2}{2} (\bar{g} + zS\cdot\zeta)^TF(g + zS\cdot\zeta) \\
     &= \frac{\eta^2}{2} (\bar{g}^T F \bar{g} + 2 \cdot zS \cdot   \zeta^TF\bar{g} + (zS)^2 \cdot \zeta^TF\zeta) \enspace . \label{eq:full_form_fisher_kl} \\
     &= a_0 + \sum_{i}b_i\zeta_i + \sum_{i,j} c_{ij} \zeta_i \zeta_j \enspace.
\end{align}

We denote $\widehat{D_\text{KL}} = \frac{\eta^2}{2} \tilde{g} F \tilde{g}$ the above random variable. It is a quadratic form in $\zeta$ which follows a generalized Chi-squared distribution (see \citet{das2024methodsintegratemultinormalscompute}, Section 2.1). %We denote $Q(\alpha)$ the quantile function of this distribution, so that $\mathbb{P}[\eta^2 \tilde{g} F \tilde{g} \ge Q(\alpha)] = 1 - \alpha$.
Below, we compute its expectation and variance.

\paragraph{Expectation.} Because $\zeta$ has zero mean, and thus $\mathbb{E}[\zeta^TF \bar{g}] = 0$:
\begin{equation*}
    \mathbb{E} \left[\widehat{D_\text{KL}} \right] = \frac{\eta^2}{2} \left(\bar{g}^T F \bar{g} + (zS)^2 \cdot\mathbb{E}[\zeta^TF\zeta]\right)
\end{equation*}

For the second term, denoting $F = (f_{ij})_{1\le i,j \le d}$, we have:
\begin{align*}
    \mathbb{E} [\zeta^TF \zeta] &= \mathbb{E} \left[\sum_{i,j} f_{ij} \zeta_i \zeta_j\right] \\
    &= \sum_{i,j} f_{ij} \mathbb{E}[\zeta_i \zeta_j] \\
    &= \sum_{i \ne j} f_{ij} \mathbb{E}[\zeta_i] \mathbb{E}[\zeta_j] + \sum_i f_{ii} \mathbb{E}[\zeta_i^2] \\
    &= 0 + \sum_{i} f_{ii} \mathbb{V}[\zeta_i] \\
    &= \text{tr}(F).
\end{align*}

Therefore, we have:
\begin{equation} \label{eq:exp_fisher_kl}
    \mathbb{E} \left[\frac{\eta^2}{2} \tilde{g}^T F \tilde{g} \right] = \frac{\eta^2}{2} \cdot \left(\bar{g}^T F \bar{g} + (zS)^2 \cdot \text{tr}(F) \right) \enspace .
\end{equation}

We also have the following bound on $\bar{g}^T F \bar{g}$:
\begin{align*}
    \bar{g}^T F \bar{g} &= \Vert F^{1/2} \bar{g} \Vert_2^2 \\
    &\le \Vert F^{1/2}\Vert_\text{op}^2 \cdot \Vert \bar{g} \Vert_2^2 \\
    &= \bar{\sigma}(F^{1/2})^2 \cdot \Vert \bar{g} \Vert_2^2 \\
    &\le \bar{\sigma}(F) \cdot S^2 \enspace ,
\end{align*}
which implies:
\begin{equation} \label{eq:exp_fisher_kl_bound}
    \mathbb{E} \left[\frac{\eta^2}{2} \tilde{g}^T F \tilde{g} \right] \le \frac{\eta^2}{2} S^2 \cdot \left(\bar\sigma(F) + z^2 \cdot \text{tr}(F) \right) = \bar{\mathbb{E}}\enspace .
\end{equation}

Using Equation (\ref{eq:exp_fisher_kl_bound}), we are now interested in deriving the value of $S$ for which the update stays within the trust region with probability $1 - \beta$. For this, we can use the Markov inequality:
$$\mathbb{P}[\widehat{D_\text{KL}} \ge \alpha] \le \frac{\mathbb{E}[\widehat{D_\text{KL}}]}{\alpha} \le \frac{\bar{\mathbb{E}}}{\alpha}$$
We want the event $[\widehat{D_\text{KL}} \ge \alpha]$ to have low probability, less than $\beta$.  To get $\bar{\mathbb{E}}/\alpha \le \beta$, we therefore have to set:
\[
    S \le \frac{1}{\eta} \sqrt{\frac{2\alpha\beta}{\bar{\sigma}(F) + z^2\text{tr}(F)}} \enspace .
\]
Then with probability at least $1 - \beta$, the noisy update states within the trust region of radius $\alpha$, \textit{i.e.}, $\widehat{D_\text{KL}} \le \alpha$.

\paragraph{Variance}. We have:
\begin{align*}
    \mathbb{V}[\widehat{D_\text{KL}}] &= \mathbb{V}[\frac{\eta^2}{2}\cdot\left(2zS \cdot \zeta^TF \bar{g} + z^2S^2 \cdot \zeta^T F \zeta\right)] \\
    &= \frac{\eta^4}{4} \left[(2zS)^2 \cdot \mathbb{V}[X] + (z^2S^2)^2 \cdot \mathbb{V} [Y] + 2 \cdot (2zS) \cdot (z^2S^2) \cdot \text{Cov}[X, Y] \right] \enspace,
\end{align*}
where $X=\zeta^TF\bar{g}$ and $Y=\zeta^T F \zeta$.

We have $\mathbb{V}[Y] = 2\text{tr}(F^2)$ (variance of a quadratic Gaussian form). Moreover:
\begin{align*}
    \mathbb{V}[X] &= \mathbb{V}\left[\sum_{i}\left(\sum_j f_{ij} \bar{g}_j \right)\zeta_i \right] \\
    &= \mathbb{V}\left[\sum_{i}h_i\zeta_i \right] \\
    &= \sum_{i}h_i^2\mathbb{V}\left[\zeta_i \right] \\
    &= \sum_i h_i^2 \cdot 1 \\
    &= \sum_{i} \left(\sum_j f_{ij} \bar{g}_j \right)^2 \\
\end{align*}

Now, $\text{Cov}[X,Y] = \mathbb{E}[(X - \mathbb{E}[X]) (Y - \mathbb{E}[Y])] = \mathbb{E}[XY - X\mathbb{E}[Y]] = \mathbb{E}[XY] - \text{tr}(F)\mathbb{E}[X] = \mathbb{E}[XY]$ because $\mathbb{E}[X] = 0$. We have:
\begin{align*}
    \mathbb{E}[XY] &= \mathbb{E}\left[\left(\sum_{i}h_i\zeta_i\right) \left(\sum_{j,k}f_{jk}\zeta_j \zeta_k\right)\right] \\
    &= \mathbb{E}\left[\sum_{i,j,k}h_i f_{jk}\zeta_i \zeta_j \zeta_k\right] \\
    &= \sum_{i,j,k}h_i f_{jk}\mathbb{E}\left[\zeta_i \zeta_j \zeta_k\right] \\
    &= 2\sum_{i,j} h_i f_{ij} \mathbb{E}[\zeta_i^2\zeta_j] + \sum_{i,j} h_i f_{jj} \mathbb{E}[\zeta_i \zeta_j^2] + \sum_{i}h_i f_{ii} \mathbb{E}[\zeta_i^3] \enspace.
\end{align*}
The first two terms are zero because $\mathbb{E}[\zeta_i] = 0$, and the third term is zero because the third moment of a zero-mean Gaussian is zero. Therefore $\text{Cov}[X, Y] = 0$. Finally:
\begin{align} \label{eq:var_kl_fisher}
    \mathbb{V}[\widehat{D_\text{KL}}] &= \frac{\eta^4}{4} \left[(2zS)^2 \cdot 2 \text{tr}(F^2) + (z^2S^2)^2 \cdot  \mathbf{1}^T F \bar{g} \right] \\
    &= \frac{\eta^4}{4} \left[8z^2S^2 \text{tr}(F^2) + z^4S^4 \cdot \sum_{i} \left(\sum_j f_{ij} \bar{g}_j \right)^2 \right] \enspace.
\end{align}
% Combining Equation~\ref{eq:var_kl_fisher} and $\mathbb{E}[\Delta \theta] = \eta^2(\bar{g}^TF\bar{g} + z^2S^2\cdot \text{tr}(F))$, and applying the Cantelli's inequality:
% \[
%     \mathbb{P}[\widehat{D_\text{KL}} - \mathbb{E}[\widehat{D_\text{KL}}] > \lambda] \le \frac{\mathbb{V}[\widehat{D_\text{KL}}]}{\mathbb{V}[\widehat{D_\text{KL}}] + \lambda}
% \]

\subsubsection{Distribution}

To derive the distributions of the trust region sizes, we rely on the results from \cite{LinearModelsTimeSeriesBook} (\href{https://onlinelibrary.wiley.com/doi/pdf/10.1002/9781119432036.app1}{Appendix A.1}). For $Y = X^TAX$ with $\mathcal{N}_d(\mu, \Sigma)$, using the spectral decomposition $\Sigma^{1/2} A \Sigma^{1/2} = P \Lambda P^T$ (with $P$ an orthogonal matrix and $\Lambda= \text{diag}(\lambda_1, ..., \lambda_d)$). Then, denoting $\nu = P^T \Sigma^{-1/2} \mu = (\nu_1, ..., \nu_d)^T$, $Y \sim \sum_{i=1}^d \lambda_i W_i^2$ with $W_i \sim \chi^2(1, \nu_i^2)$, where $\chi^2(1, \nu_i^2)$ is a non-central $\chi^2$ distribution with one degree of freedom and non-centrality parameter $\lambda_i$. 

We start with the more intricate result involving the Fisher distribution before addressing the case of standard PG, where the distribution actually takes a particularly simple form. 

\paragraph{TRPO Trust Region}

The (random) trust region size can be written as:
\[
    \frac{1}{2}\Delta \theta^T F \Delta \theta = X^T A X \enspace,
\]
with $X = \frac{\eta}{\sqrt{2}}(\bar{g} + \xi) \sim \mathcal{N}_d(\mu, \Sigma)$ ($\mu = \frac{\eta}{\sqrt{2}} \bar{g}$ and $\Sigma = \frac{\eta^2}{2} z^2S^2 I_d$) and $A = F$. In particular, its marginals $X_i$'s are i.i.d. with $X_i = \mathcal{N}(\mu_i, \sigma^2)$. with $\mu_i = \eta\bar{g}_i$ and $\sigma^2=\eta^2 z^2S^2$.  With these notations, we can apply the results from \citet{LinearModelsTimeSeriesBook}.

We have $\Sigma^{1/2} A \Sigma^{1/2} = \frac{\eta^2}{2} z^2 S^2 F = P (\frac{\eta^2}{2} z^2 S^2 \text{diag}(\sigma_1(F), ..., \sigma_d(F))) P^T$, where the $\sigma_i(F)$ are the eigenvalues/singular values of $F$. Then $\nu = P^T \cdot (\frac{\eta}{\sqrt{2}} zS)^{-1} I_d \cdot \frac{\eta}{\sqrt{2}} \bar{g} = \frac{1}{zS} P^T \bar{g}$. Therefore:
\begin{equation}
    \frac{1}{2}\Delta\theta^T F \Delta \theta \sim \frac{\eta^2}{2}z^2S^2 \sum_{i=1}^d \sigma_i(F) \chi^2(1, \nu_i^2), \enspace.
\end{equation}

\paragraph{Standard Policy Gradient Trust Region}

The (random) trust region size can be written as:
\[
    \frac{\Delta \theta^T\Delta \theta}{2} = X^T X = \sum_{i=1}^d X_i^2\enspace ,
\]
with $X = \frac{\eta}{\sqrt{2}}(\bar{g} + \xi)$. We have $X \sim \mathcal{N}_d(\frac{\eta}{\sqrt{2}}\bar{g}, \frac{\eta^2}{2}z^2S^2 I_d)$. In particular, its marginals $X_i$'s are i.i.d. with $X_i = \mathcal{N}(\mu_i, \sigma^2)$. with $\mu_i = \frac{\eta}{\sqrt{2}}\bar{g}_i$ and $\sigma^2=\frac{\eta^2}{2}z^2S^2$. Because $X_i = \frac{\eta}{\sqrt{2}}zS \cdot \mathcal{N}(\frac{\bar{g}_i}{zS}, 1)$, we can reformulate this in terms of unit-variance normal variables $Z_i \sim \mathcal{N}(\frac{\bar{g}_i}{zS}, 1)$:
\[
    \frac{\Delta \theta^T\Delta \theta}{2} = \frac{\eta^2}{2} z^2 S^2 \sum_{i=1}^d Z_i^2 \enspace.
\]
We can then apply the following result: if $(Z_1, ..., Z_d)$ are $d$ independent, normally distributed with mean $\mu_i$ and unit variance, then $\sum_{i=1}^d Z_i^2$ follow a non-central $\chi^2$ distribution with $d$ degrees of freedom and non-centrality parameter $\lambda = \sum_{i=1}^d \mu_i^2$, denoted $\chi^2(d, \lambda)$.

In our case, the non-centrality parameter is:
\[
    \lambda = \sum_{i=1}^d \frac{\bar{g}_i^2}{z^2S^2} = \frac{\Vert \bar{g} \Vert^2}{z^2S^2}  \enspace,
\]
and therefore:
\[
    Z := \sum_{i=1}^d Z_i^2 \sim \chi^2 \left(d, \frac{\Vert \bar{g} \Vert^2}{z^2S^2} \right) \enspace.
\]

We can numerically compute the PDF, CDF and PPF of $Z$ using, for instance, \texttt{scipy.stats.ncx2} in Python. In particular, for $\beta \in (0,1)$, we can compute the quantile $q_{1-\beta}$ such that $\mathbb{P}\left(Z \le q^{\chi^2(d, \Vert\bar{g}\Vert^2/z^{-2}S^{-2})}_{1-\beta}\right) = 1- \beta$.

Therefore, let us assume that we want the update to remain within a $L_2$ trust region of size $\alpha$ with probability $1 - \beta, \textit{i.e.}$, $\mathbb{P}(\frac{\Delta \theta^T \Delta \theta}{2} \le \alpha) = 1 - \beta$. We have:
\begin{align*}
    \mathbb{P}(\frac{\Delta \theta^T \Delta \theta}{2} \le \alpha) &= \mathbb{P}( \frac{\eta^2}{2} z^2 S^2 \cdot Z\le \alpha) \\
    &= \mathbb{P}(Z\le \frac{2 \alpha}{\eta^2 z^2S^2}) .
\end{align*}

From the definition of the quantile function, $\mathbb{P}(Z \le q^{\chi^2(d, \Vert\bar{g}\Vert^2/z^{-2}S^{-2})}_{1-\beta}) = 1 - \beta$. To derive an inequality involving $S$ and the other parameters of the problem, we bound $\frac{\Vert \bar{g} \Vert^2}{z^{2}S^{2}} \le \frac{1}{z^2}$ and use the fact that the function of $\lambda \mapsto q^{\chi^2(d, \lambda)}_x$ is increasing in $\lambda$: $q^{\chi^2(d, \Vert\bar{g}\Vert^2/z^{-2}S^{-2})}_{1-\beta} \le q^{\chi^2(d, z^{-2})}_{1-\beta}$. Therefore $\mathbb{P}(Z \le q^{\chi^2(d, \Vert\bar{g}\Vert^2/z^{-2}S^{-2})}_{1-\beta}) \le \mathbb{P}(Z \le q^{\chi^2(d, z^{-2})}_{1-\beta})$, and it follows that $\mathbb{P}(Z \le q^{\chi^2(d, z^{-2})}_{1-\beta}) \ge 1-\beta$.

Equalizing $\frac{2 \alpha}{\eta^2 z^2S^2} = q^{\chi^2(d, z^{-2})}_{1-\beta}$ and deriving $S$, we have:
\[
    S \le \frac{1}{\eta z} \sqrt{\frac{2 \alpha}{q^{\chi^2(d, z^{-2})}_{1-\beta}}} \enspace.
\]
Therefore, setting $S$ like above, we have $\mathbb{P}(\frac{\Delta \theta^T \Delta \theta}{2} \le \alpha) \ge 1 - \beta$.
\newpage
\section{Additional discussions}

\subsection{Computing the Privacy Guarantees for Algorithm~\ref{alg:dp_pg}} \label{sec:app_priv_guarantees_dp_pg}

Given $\delta \in (0,1)$, for $\epsilon \in (0, 1)$, the Gaussian mechanism $\mathcal{M}_1$ with scale $\sigma = \frac{\sqrt{2 \log(1.25/\delta)}S}{\epsilon}$ guarantees $(\epsilon, \delta)$-DP. Therefore, for a noise multiplier $z > \sqrt{2 \ln(1.25/\delta)} := C_1(\delta)$, the privacy budget is $\epsilon = \frac{C_1(\delta)}{z}$. For instance, $C(10^{-2}) \approx 3.1$ and $C_1(10^{-5}) \approx 4.8$.  We stress that this mechanism only works for $\epsilon < 1$, which is often overlooked in the literature, leading to incorrect evaluation of the privacy guarantees. Given $\delta \in (0,1)$ and a noise multiplier $z$, we can only compute $\epsilon = \frac{C_1(\delta)}{z}$ if $z > C_1(\delta)$ with the above mechanism. For smaller values of $z$, this formula returns a value $\epsilon \ge 1$ and the above mechanism no longer holds.

Fortunately, \citet{improved_gaussian_mechanisms} provide Gaussian mechanisms for the case $\epsilon \ge 1$: according to Theorem 5, given $\delta \in (0, 5)$ and $C_2(\delta)=\sqrt{\log{\frac{2}{\sqrt{16 \delta + 1} - 1}}}$, the Gaussian mechanism $\mathcal{M}_2$ with scale $\sigma = \frac{(C_2(\delta) + \sqrt{C_2(\delta)^2 + \epsilon}) \cdot \Delta}{\epsilon \sqrt{2}}$ is $(\epsilon, \delta)$-DP. Inverting the relationship, we can derive $\epsilon$ given $\delta$ and $z$ as $\epsilon = \frac{1 + 2\sqrt{2}C_2(\delta)z}{2 z^2}$. Numerically, we observe that $\mathcal{M}_2$ yields slightly worse $\epsilon$ than $\mathcal{M}_1$ for the same $z$. For instance, given $\delta = 10^{-5}$, $\mathcal{M}_1$ yields $\epsilon \approx 4.8$ while $\mathcal{M}_2$ yields $\epsilon \approx 5.0$. The latter is the correct value of the privacy budget $\epsilon$.

\begin{figure}[ht]
\vskip 0.2in
\begin{center}
\centerline{\includegraphics[width=0.5\columnwidth]{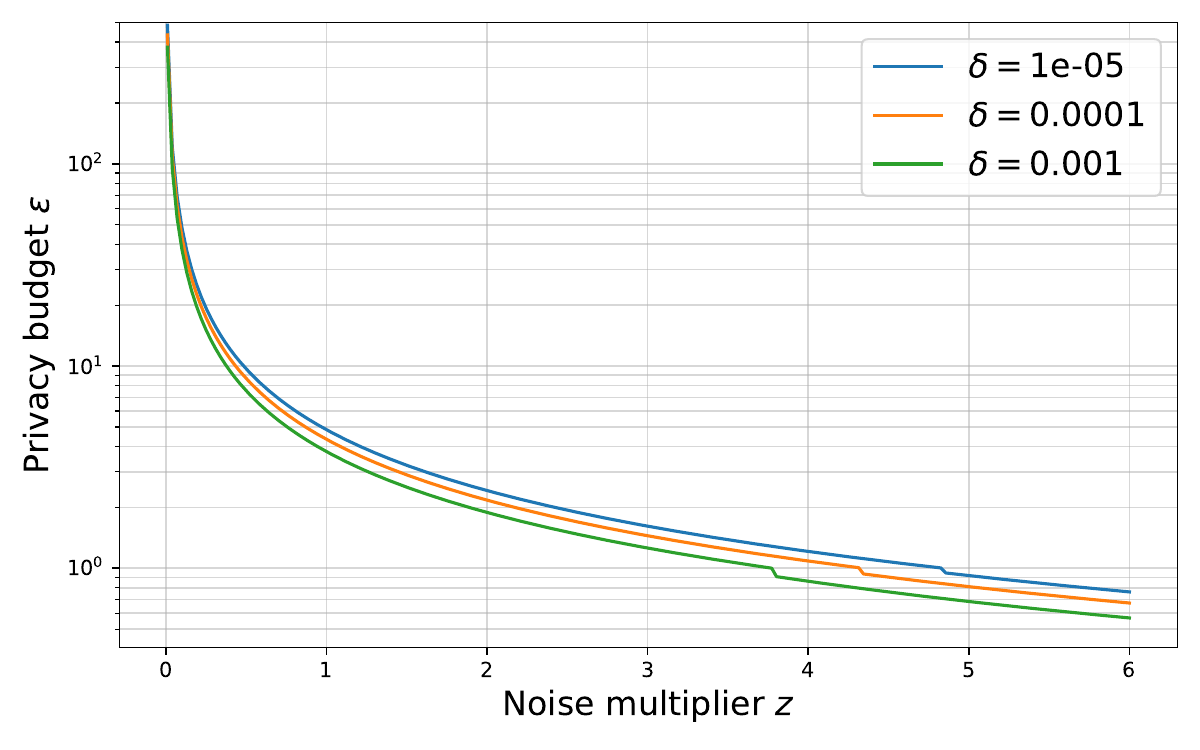}}
\caption{Relationship between the noise multiplier $z$ ($x$\textit{-axis}) and the privacy budget $\epsilon$ ($y$\textit{-axis}) of Algorithm~\ref{alg:dp_pg}.}
\label{fig:z_eps_relation}
\end{center}
\vskip -0.2in
\end{figure}

Figure~\ref{fig:z_eps_relation} plots the relationship between $z$ and the total privacy budget $\epsilon$ spent by Algorithm~\ref{alg:dp_pg}. We can a break in the curves at $\epsilon = 1$, corresponding to the switch between mechanisms $\mathcal{M}_1$ (right portion of the curves) and $\mathcal{M}_2$ (left portion of the curves). We can see that $\mathcal{M}_2$ provides less optimistic privacy budgets (larger $\epsilon$) that $\mathcal{M}_1$ would if we (incorrectly) applied the latter for smaller values of $z$, although the difference is rather small.

\section{Experiment details} \label{sec:app_exps_details}

\subsection{Task details}

\begin{figure}
    \centering
    \includegraphics[width=0.8\linewidth]{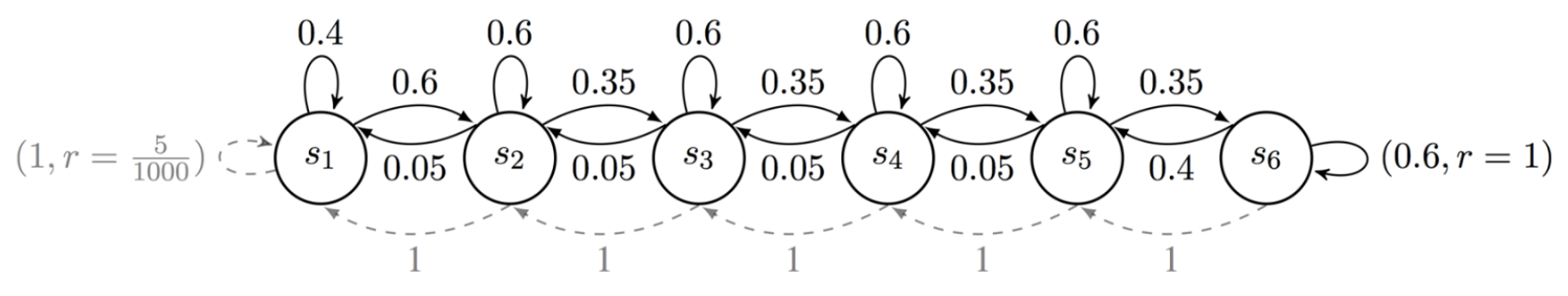}
    \caption{The Riverswim environment (taken from \cite{chowdhury_differentially_2021}).}
    \label{fig:riverswim_env}
\end{figure}

\paragraph{Riverswim.} \textsc{Riverswim} \cite{Riverswim_OsbandRR13} is an episodic tabular MDP with horizon $H=20$, 6 states and 2 actions (\textit{swim left} and \textit{swim right}). The agent starts from the left bank of the river, and can get a small reward $r = 5/1000$ by swimming left toward the bank. However, the agent can get a much larger reward $r=1$ by swimming up the river, reaching the right bank, and swim right towards the right bank. This MDP thus requires some amount of exploration. Due to the river current, the swimmer may choose the action \textit{swim right} and end up staying in their current state, or even move left, making the environment stochastic. Figure~\ref{fig:riverswim_env} illustrates the environment. The transition probabilities influence the difficulty of the task, and we try different probabilities of swimming right and obtaining the rewards at the rightmost state.

For our log-linear policy, we use as a feature a concatenation of the one-hot encoded state and the action.

\paragraph{Dosing.} To better illustrate the real-world relevance of our approach, we learn DP personalized medication policies trained on simulated patient data. In the \textsc{Dosing} task, the goal is to keep blood glucose at tolerable levels in Type-1 diabetic patients, and the RL agent controls the insulin uptake. Each trajectory corresponds to a full day of patient measures, and the state contains various health measures such as meal disturbance amounts, insulin levels and blood glucose levels, which can be considered as sensitive and private information. 
We use two versions of the environment: \textsc{Dosing-0}, where the meal pattern is constant across episodes, and the more difficult \textsc{Dosing-1} where each patient (or episode) has a different meal pattern throughout the day. The simulator is presented in \href{https://www.strong.io/blog/reinforcement-learning-for-personalized-medication-dosing}{this blog post}, and we borrow the official implementation from \href{https://github.com/strongio/dosing-rl-gym}{this GitHub repository}.

\paragraph{Sentiment fine-tuning with RLHF.} We start from the following Huggingface's TRL illustrating example: \url{https://huggingface.co/docs/trl/en/sentiment_tuning}. The goal is to fine-tune a GPT-2 model with 137M parameters on an IMDB review dataset to encourage the model to produce a positive review. As a reward model, we use a BERT sentiment classifier. Precisely, given a prompt $s_t$ (the beginning of a review), the GPT-2 model $\pi_\theta$ completes the review with a sequence $a_t \sim \pi_\theta(\cdot \vert s_t)$. The reward model $R$ gives a score $r_t = R([s_t, a_t])$ to the resulting review, such that $r_t>0$ is the review is positive, and $r_t<0$ otherwise. A policy gradient algorithm is used to optimize the policy parameters $\theta$ so that the model achieves larger rewards during training. In the non-private case, we use the PPO algorithm, and use our \text{DPPG} algorithm with $z=0.01$ in the private case. As a starting point, we use a GPT-2 model already fine-tuned on IMDB data (\url{https://huggingface.co/lvwerra/gpt2-imdb}).

\subsection{Implementation details}

\subsubsection{General}

\paragraph{Actor-Critic style.} Algorithm~\ref{alg:dp_pg} is also compatible with the actor-critic framework where we also train a value function $V_\psi$ to compute the advantage estimates. According to our threat model, the adversary observes the policy $\pi_\theta$, and therefore we only need to privatize the policy updates. If however some parameters are shared between $\pi_\theta$ and $V_\psi$ (often the case with visual inputs), it becomes necessary to make the value function updates private as well, and we proceed as for the policy updates.

\subsubsection{Log-linear \textsc{DPPG}}

For the tabular MDP \textsc{Riverswim}, we use a log-linear policy:
\[
    \pi_\theta(a \vert s) \propto \exp(\theta^T \phi_{s,a})  \enspace,
\]
and a linear baseline $V_\psi(s) = \Phi^T\phi_{s,a}$. We use a the following learning rate scheme for the policy: we start from $\eta = \eta_0$, and every $E_\eta$, update the learning rate with $\eta \leftarrow \eta / k_\eta$ with $k_\eta > 0$.

\subsubsection{Deep \textsc{DPPG}}

Implementation of Algorithm~\ref{alg:dp_pg} is based on the \href{https://github.com/vwxyzjn/cleanrl}{CleanRL implementation} of \textsc{PPO}. 

\paragraph{Private adaptive optimization.} To help convergence, we use the \textsc{Adam} optimizer during the local update step (\textit{i.e.}, within the \textsc{ComputeLocalUpdate} function). Adam keeps track of the two first moments of the gradient to adapt the learning rate at each epoch. In the private setting, we must be particularly careful not to pass noise-free gradient information from one user to the next through the optimizer. Indeed, if we would use the non-private gradient of user $k$ to update Adam's estimates of the moments, this information would be used in the local update of another use $\ell$ at the next iteration. Therefore, we would actually use several times the gradient information of user $k$, resulting in more privacy leakage due to the sequential composition property of DP. To avoid this, at the start of each iteration, we substitute the Adam internal statistics with the private global gradient $\tilde{g}$ and its square $\tilde{g}^2$. Adam is then used as usual during the local update phaase, since the data of only one user is concerned.

\subsection{Hyperparameters}

Tables \ref{tab:training_hyper_tabular} and \ref{tab:training_hyper} present the hyperparameters used to obtain the results from Section~\ref{sec:exps}.

\begin{table}[ht]
\vskip 0.15in
\begin{center}
\begin{small}
\begin{sc}
\begin{tabular}{lccc}
\toprule
 & Riverswim \\
 \midrule
Discount factor $\gamma$ & 0.99 \\
Start policy LR $\eta_0$ & 12  \\
Policy LR update freq. $E_\eta$ & 50 ep. \\
Policy LR update factor $k_\eta$ & 5.0 \\
Min. policy LR & 0.06 \\
\midrule
Trust region size $\alpha$ & 3.5 \\
Confidence $1 - \beta$ & 0.6 \\
\midrule
Num. Fisher episodes & 25 \\
Fisher regularizer & $1.10^{-3}$ \\
\bottomrule
\end{tabular}
\end{sc}
\end{small}
\end{center}
\caption{Training and Hyperparameters details for Tabular MDP.}
\label{tab:training_hyper_tabular}
\vskip -0.1in
\end{table}

\begin{table}[ht]
\vskip 0.15in
\begin{center}
\begin{small}
\begin{sc}
\begin{tabular}{lccc}
\toprule
 & Control & MuJoCo & Dosing \\
 \midrule
Discount factor $\gamma$ & 0.99 & 0.99 & 0.99 \\
Learning rate $\eta$ & $7.26 \times 10^{-4}$ & $2.04 \times  10^{-4}$ &  $9.25 \times  10^{-4}$ \\
Num. local minibatch's & 2 & 64 & 2 \\
Local epochs & 8 & 8 & 4 \\
Num. steps $T$ & 64 & 2048 & 64 \\
Clipping norm $S$ & 0.05 & 1.80 &  0.08\\
Entropy coef. & 0.36 & 0.02 & 0.01 \\
GAE lambda & 0.85 & 0.91 & 0.97 \\
Num. hidden neurons & 64 & 64 & 64 \\
Users per update $K$ & 8 & 8 & 8 \\
\bottomrule
\end{tabular}
\end{sc}
\end{small}
\end{center}
\caption{Training and Hyperparameters details.}
\label{tab:training_hyper}
\vskip -0.1in
\end{table}
\newpage
\section{Tables}

\begin{table}[ht]
\caption{Examples of sensitive information that could be revealed through the states and rewards for different application scenarios of personalized RL.}
\label{tab:ex_sensitive_info}
\vskip 0.15in
\begin{center}
\begin{small}
\begin{sc}
\begin{tabular}{lcc}
\toprule
Application & States & Rewards \\
\midrule
Treatment Recommendation & Patient's medical history & Patient's response to treatment \\
Personalized Finance & User's financial record & User's financial transactions \\
Personalized Advertisement  & User's browsing history & User's preferences (clicked ads) \\
\bottomrule
\end{tabular}
\end{sc}
\end{small}
\end{center}
\vskip -0.1in
\end{table}

% \begin{table}[t]
% \caption{Clipping norms $S$ for which noisy updates stay within the trust region of radius $\alpha$ in expectation, \textit{i.e.}, $\mathbb{E}[\Delta\theta^T F \Delta\theta] \le \alpha$, for different mechanisms. Each mechanism guarantees the same level of privacy.}
% \label{tab:clip_formulas}
% \vskip 0.15in
% \begin{center}
% \begin{small}
% \begin{sc}
% \begin{tabular}{cccc}
% \toprule
% Gradient & Clipping & Noise & $S$ \\
% \midrule
% Standard gradient & $\Vert g \Vert_2 \le S$ & $\zeta \sim \mathcal{N}(0, z^2S^2 \cdot I_d)$ & $S = \frac{1}{\eta} \sqrt{\frac{\alpha}{\bar{\sigma}(F) + z^2\text{tr}(F)}}$ \\
% \midrule
% Standard gradient & $\Vert g \Vert_F \le S$ & $\zeta \sim \mathcal{N}(0, z^2 S^2 F^{-1})$ & $S = \frac{1}{\alpha} \sqrt{\frac{\delta}{1 + dz^2}}$\\
% \midrule
% Natural gradient  & $\Vert F^{-1}g \Vert_2 \le S$ & $\zeta \sim \mathcal{N}(0, z^2S^2 \cdot I_d)$ & $S = \frac{1}{\eta} \sqrt{\frac{\alpha}{\frac{1}{\bar{\sigma}(F)} + z^2\text{tr}(F)}}$ \\
% \bottomrule
% \end{tabular}
% \end{sc}
% \end{small}
% \end{center}
% \vskip -0.1in
% \end{table}
\newpage
\section{Algorithm} \label{sec:app_algos}

\begin{algorithm}[ht]
   \caption{Compute Local Update - PPO}
   \label{alg:local_grad_ppo}
\begin{algorithmic}
   \STATE {\bfseries Input:} Current policy parameters $\theta$, data $\tau_u$, advantages estimates $\hat{A}_u$, clipping norm $S$, local epochs $E$, minibatch size $B$, 
   \STATE Copy current parameters $\theta_0 \leftarrow \theta$.
    \FOR{$e=1,...,E$}
        \STATE $\mathcal{B}_u \leftarrow$ ($u$'s data split into size $B$ batches)
        \FOR{batch $b \in \mathcal{B}_u$}
            \STATE $L(\theta; b) \leftarrow B^{-1}\sum_{(s^u_t, a^u_t, r^u_t) \in b} \frac{\pi_\theta(a^u_t \vert s^u_t)}{\pi_{\theta_0}(a^u_t \vert s^u_t)} \hat{A}^u_t$
            \STATE $\theta \leftarrow \theta - \eta \nabla L(\theta; b)$
            \STATE $\theta \leftarrow \theta_0 + \text{Clip}(\theta - \theta_0; C)$
        \ENDFOR
    \ENDFOR
    \STATE Return update $g_u = \theta - \theta_0$
\end{algorithmic}
\end{algorithm}
\newpage
\section{Figures}

\begin{figure}[ht]
\centering
\begin{minipage}{.49\textwidth}
%\vspace{0pt}
\includegraphics[width=\linewidth]{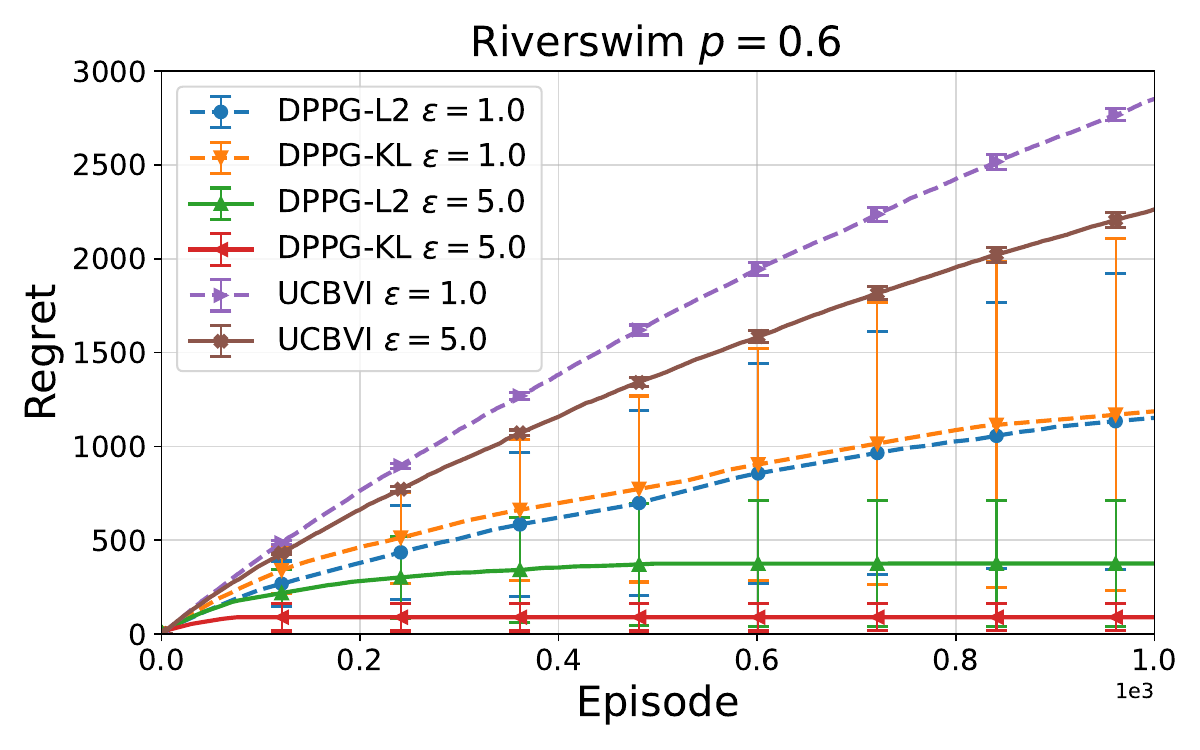}
\end{minipage}
\begin{minipage}{.49\textwidth}
%\vspace{0pt}
\includegraphics[width=\linewidth]{plots/plot_results_riverswim_9.pdf}
\end{minipage}
\caption{Cumulative regret on \textsc{Riverswim} for $\epsilon=1.0$ (\textit{dashed line}) and $\epsilon=5.0$ (\textit{solid line}).}
\label{fig:riverswim_lin}
\end{figure}

\begin{figure*}[t]
\centering
\begin{minipage}{.49\textwidth}
%\vspace{0pt}
\includegraphics[width=\linewidth]{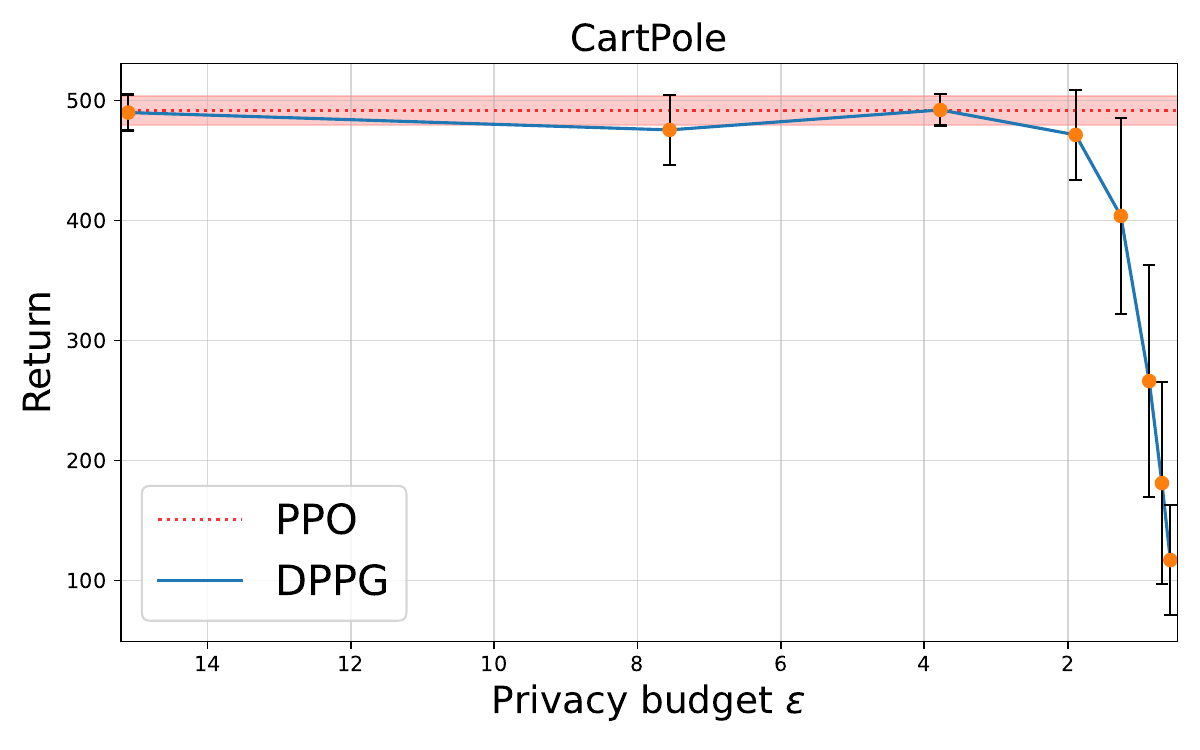}
\end{minipage}
\begin{minipage}{.49\textwidth}
%\vspace{0pt}
\includegraphics[width=\linewidth]{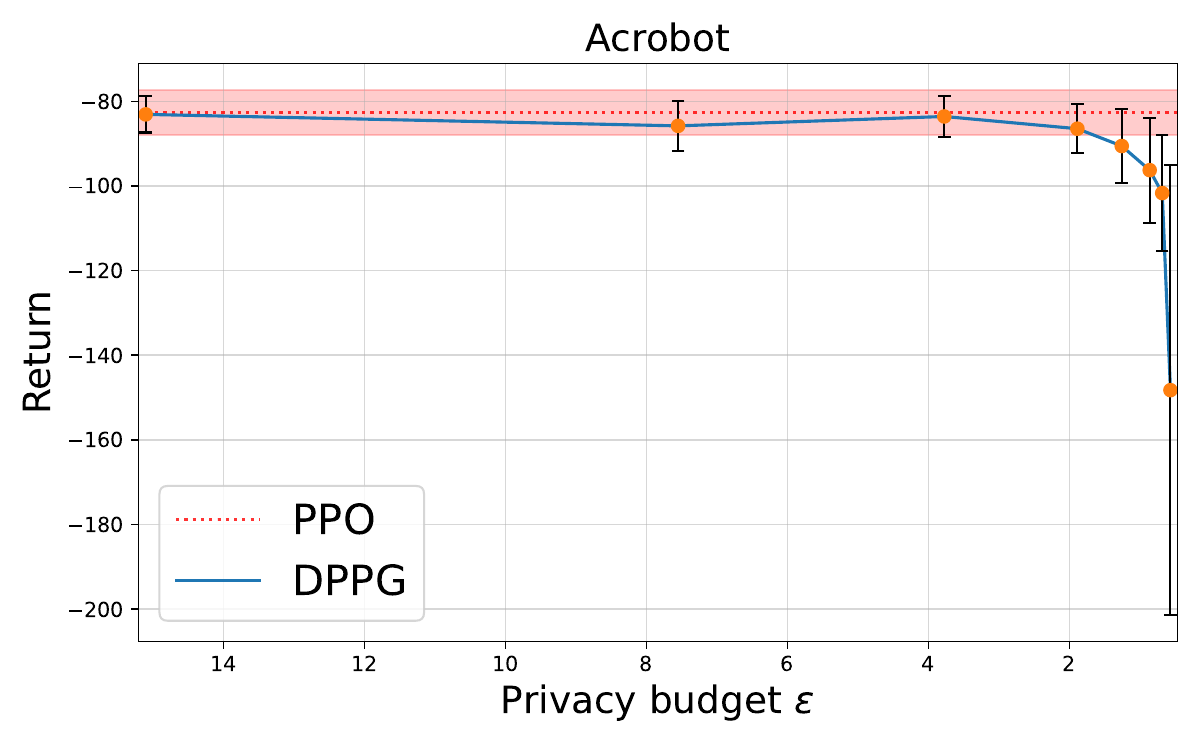}
\end{minipage}
%\hfill
%\vspace{-2pt}
\caption{Asymptotic performance vs. privacy budget $\epsilon$ \textsc{CartPole} (\textit{left}) and \textsc{Acrobot} (\textit{right}) for noise multipliers $z \in \{0.25, 0.5, 1.0, 2.0, 3.0, 4.0, 5.0, 6.0\}$.}
\label{fig:control_f_eps_lin}
\end{figure*}

\begin{figure*}[ht]
\centering
\includegraphics[width=0.8\linewidth]{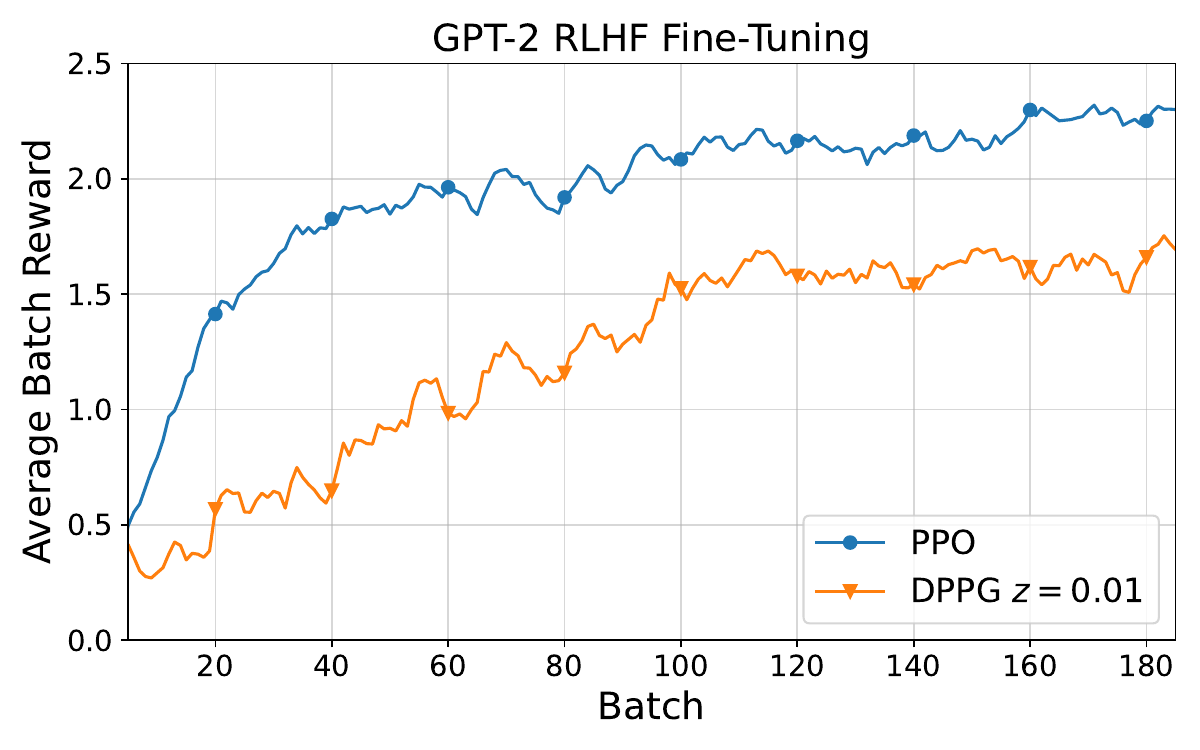}
\caption{Learning curves on the RLHF sentiment tuning task.}
\label{fig:gpt2}
\end{figure*}

\begin{figure*}[ht]
\centering
\begin{minipage}{.49\textwidth}
%\vspace{0pt}
\includegraphics[width=\linewidth]{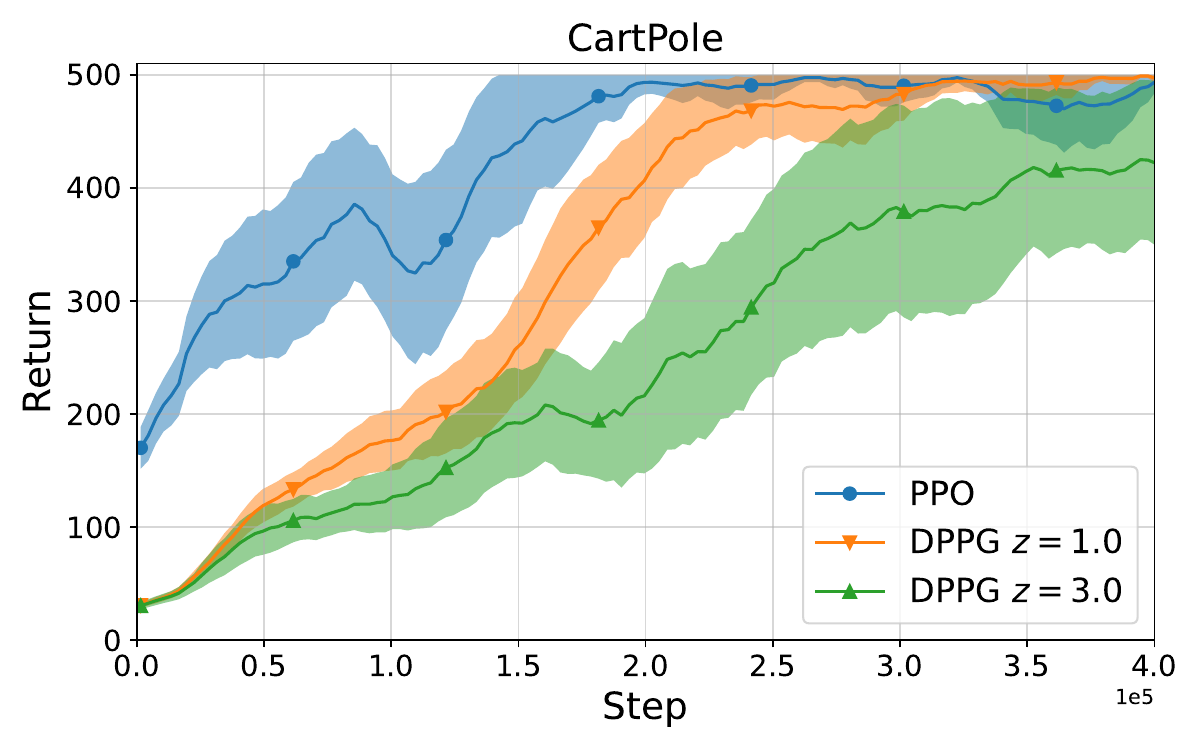}
\end{minipage}
\begin{minipage}{.49\textwidth}
%\vspace{0pt}
\includegraphics[width=\linewidth]{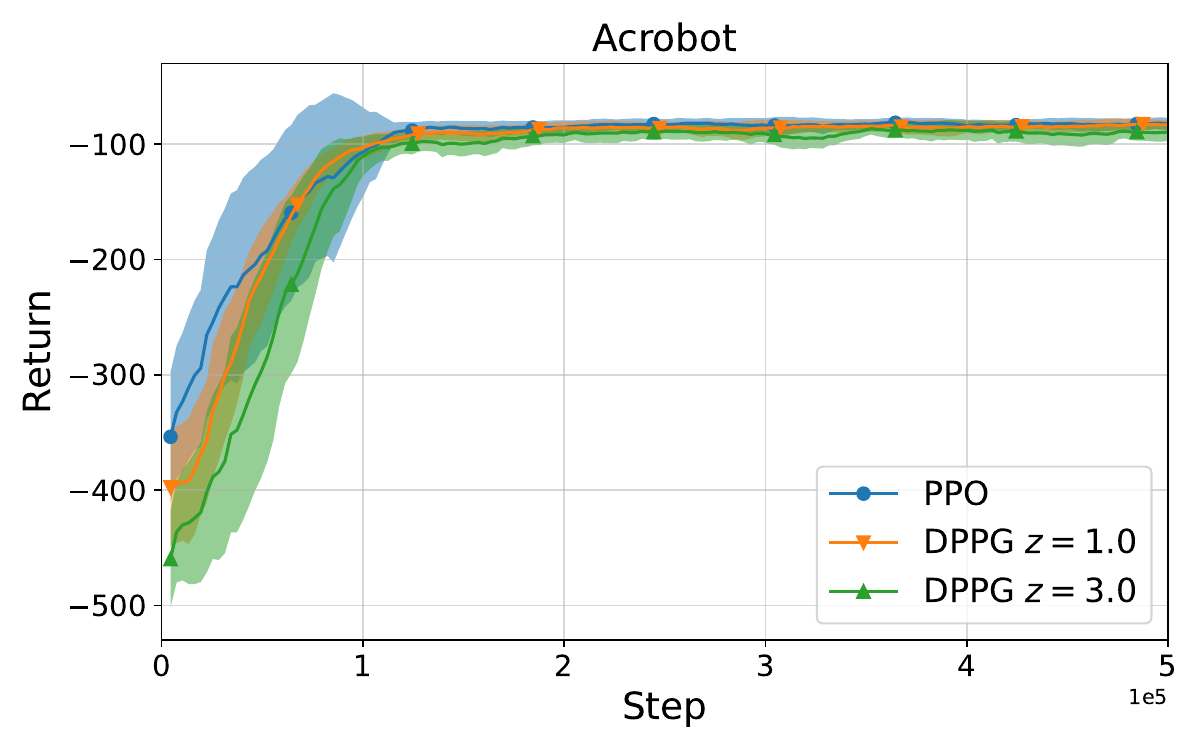}
\end{minipage}
%\hfill
%\vspace{-2pt}
\begin{minipage}{.49\textwidth}
%\vspace{0pt}
\includegraphics[width=\linewidth]{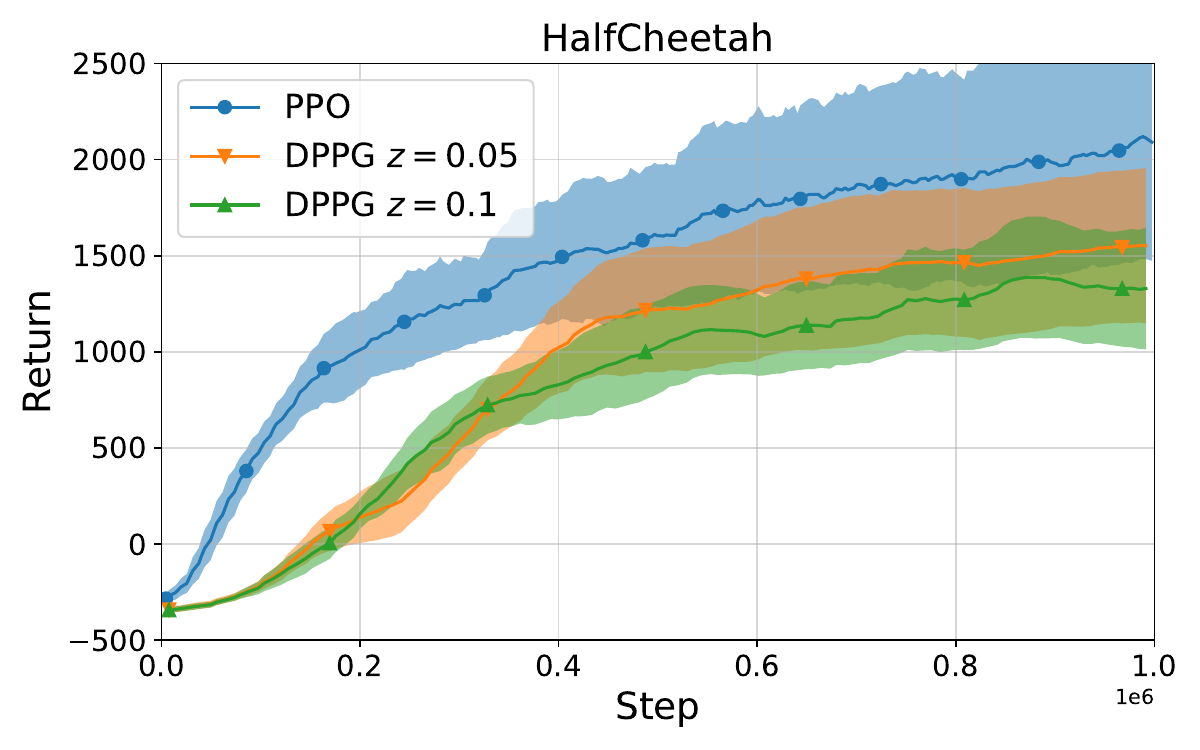}
\end{minipage}
\begin{minipage}{.49\textwidth}
%\vspace{0pt}
\includegraphics[width=\linewidth]{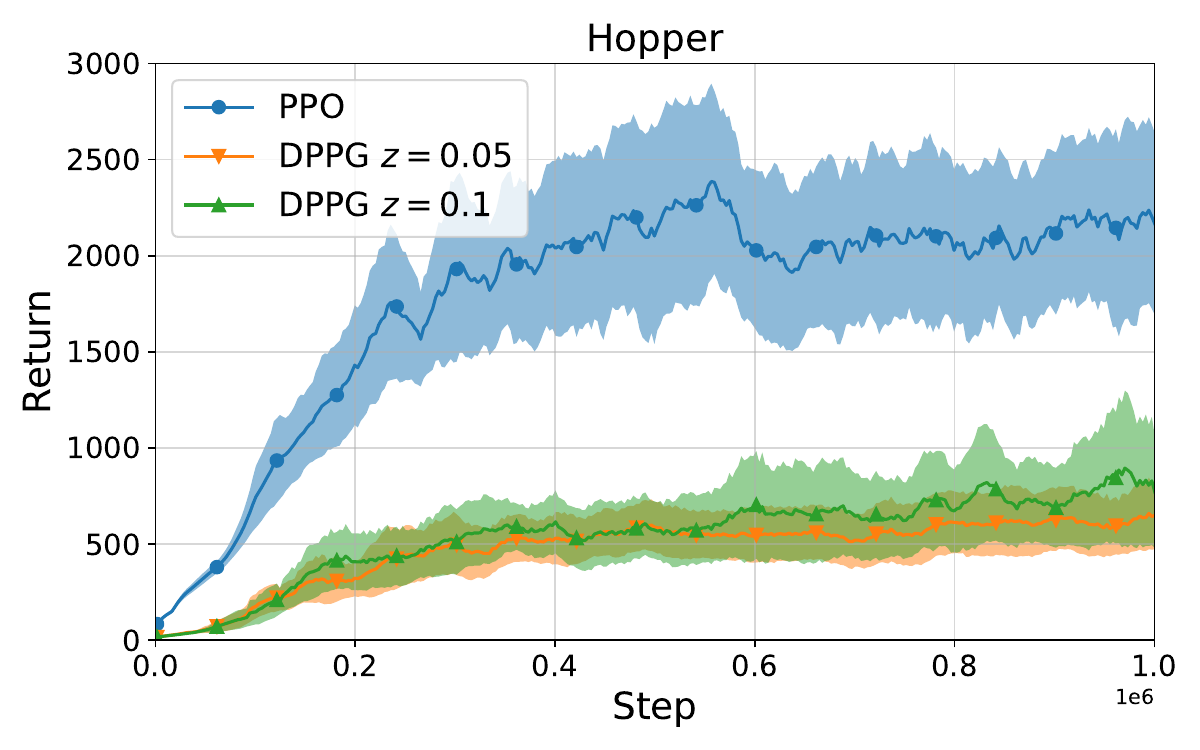}
\end{minipage}
%\hfill
%\vspace{-2pt}
\begin{minipage}{.49\textwidth}
%\vspace{0pt}
\includegraphics[width=\linewidth]{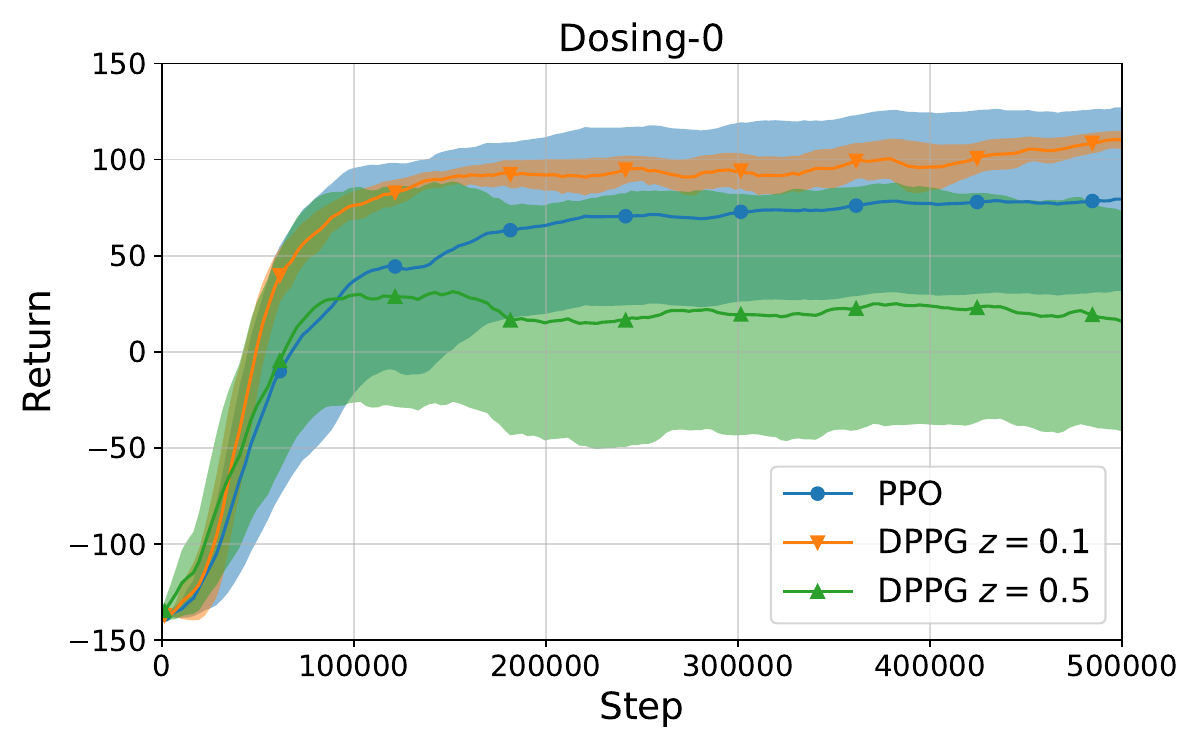}
\end{minipage}
\begin{minipage}{.49\textwidth}
%\vspace{0pt}
\includegraphics[width=\linewidth]{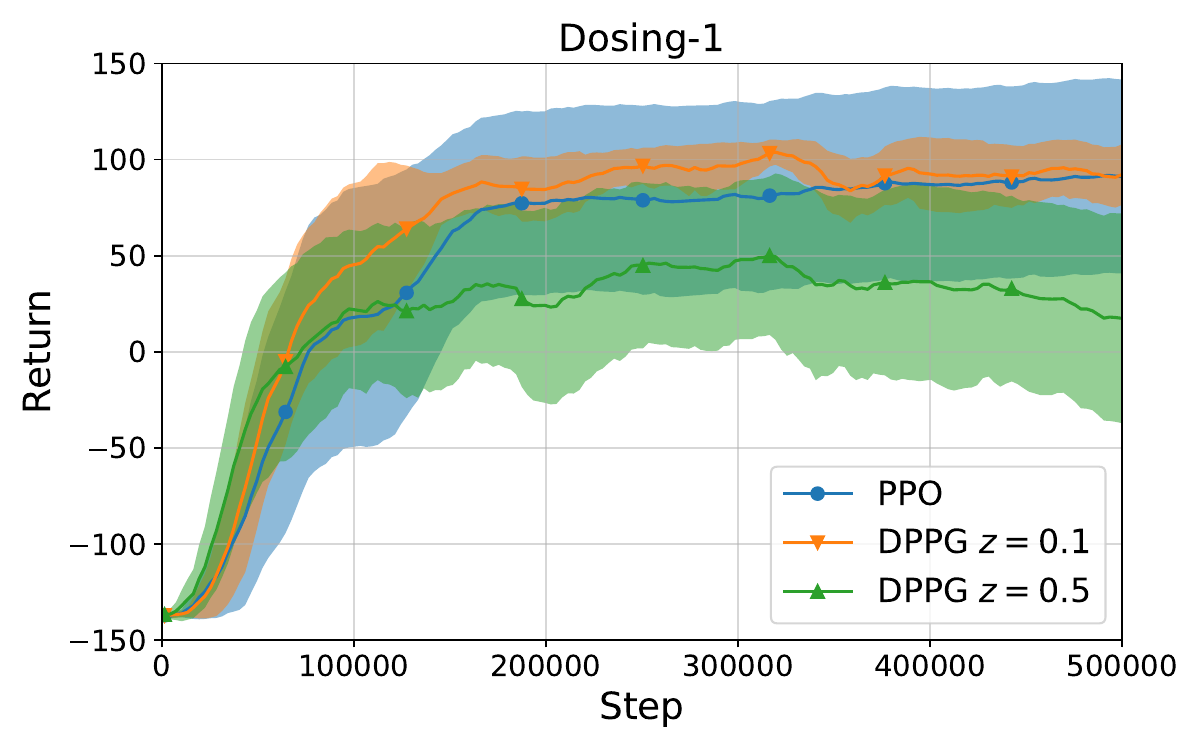}
\end{minipage}
%\hfill
%\vspace{-2pt}
\caption{Learning curves on \textsc{Control} (\textit{top}), \textsc{MuJoCo} (\textit{middle}) and \textsc{Dosing} (\textit{bottom}) environments.}
\label{fig:all_plots}
\end{figure*}

%%%%%%%%%%%%%%%%%%%%%%%%%%%%%%%%%%%%%%%%%%%%%%%%%%%%%%%%%%%%%%%%%%%%%%%%%%%%%%%
%%%%%%%%%%%%%%%%%%%%%%%%%%%%%%%%%%%%%%%%%%%%%%%%%%%%%%%%%%%%%%%%%%%%%%%%%%%%%%%

\end{document}